%% file: arxiv.tex
\documentclass[letterpaper]{article} 
\usepackage{arxiv/arxiv}  

\pdfoutput=1

\usepackage[hyphens]{url}  
\usepackage{graphicx} 
\usepackage{natbib}
\usepackage{caption}
\usepackage{algorithm}
\usepackage{algorithmic}

\usepackage{newfloat}
\usepackage{listings}
\DeclareCaptionStyle{ruled}{labelfont=normalfont,labelsep=colon,strut=off}
\floatstyle{ruled}
\newfloat{listing}{tb}{lst}{}
\floatname{listing}{Listing}

\usepackage{booktabs}

\usepackage{amsmath}
\usepackage{amssymb}
\usepackage{mathtools}
\usepackage{enumitem}
\usepackage{tabularx}
\usepackage{minibox}
\usepackage[table]{xcolor}
\usepackage{array}
\usepackage{makecell}
\usepackage{multirow}
\usepackage{dsfont}

\newcommand{\rng}[2]{#1 & #2}
\nocopyright
\newcommand{\mathbbm}[1]{\mathds{#1}}

\title{One Adapter Pair per Model: \\A Universal Activation Interface for Language Models}

\author{
    Su-Hyeon Kim\textsuperscript{\rm 1},
    Jiwan Mun\textsuperscript{\rm 2},
    Yo-Sub Han\textsuperscript{\rm 2}\corresponding
}
\affiliations{
    \textsuperscript{\rm 1}Department of Artificial Intelligence, Yonsei University\\
    \textsuperscript{\rm 2}Department of Computer Science, Yonsei University\\
    Seoul, South Korea\\
    \{suhyeon.kim, munjw7055, emmous\}@yonsei.ac.kr
}

\begin{document}

\maketitle

\begin{abstract}
Activation-based tools are usually tied to one model's native hidden
space, requiring probes, sparse autoencoders, and natural-language interpreters to be rebuilt or rediscovered for
each new language model.
We present a \emph{Universal Activation Bus}, a framework that provides
a common activation interface across compatible language models.
Using a small set of source models, we learn a shared dense space
together with one lightweight linear encoder--decoder adapter pair per model.
After source training, the interface is frozen; a new model joins by
fitting only its adapter pair on unlabeled matched text.
The resulting interface allows activation-based tools to be shared
across connected models, including common probes and SAE features as
well as access to an NLA originally trained for a different model.
Across five models, semantically related texts form consistent
neighborhoods in the shared space, and an onboarded model reuses these
tools effectively without retraining them.
We further show that an intermediate activation from one model can be
used by another model's frozen upper layers to produce predictions.
These results establish a stable, model-wise activation contract for
reusable tools across compatible language models.
Code is available at public repository\footnote{\url{https://github.com/suhyeon0123/Universal_Activation_Bus}}.
\end{abstract}

\section{Introduction}

Understanding and manipulating the hidden states of large language models (LLMs) is important for interpretability, control, and reliable deployment.
A growing class of \emph{activation-based methods} operates directly on hidden states: sparse autoencoders (SAEs) expose reusable features, natural-language autoencoders (NLAs) describe activations in text, and activation steering modifies generation without updating model weights~\citep{gao2024scaling,anthropic2026nla,panickssery2024caa,oozeer2025interventions}.
Yet these methods and the representations they learn are usually defined in one model's native hidden space.
As LLM families diversify in hidden dimension, tokenizer, and training distribution, the same tools must be rebuilt for each model.
This raises a practical question:
\emph{when a new model is released, can it reuse an existing ecosystem of activation-based tools through a shared interface?}

Prior work provides partial evidence that this may be possible.
Representation-similarity studies find related geometric structure
across independently trained models~\citep{kornblith2019cka,huh2024platonic}.
Relative coordinates, learned translators, anchor projections, and
feature-alignment methods go further by constructing comparable
representations or transferring selected tools across
models~\citep{moschella2023relative,jha2025vec2vec,kim2026acs,chen2025featurestitching,oozeer2025interventions}.
However, these approaches typically build a space for one tool or task,
or learn connections for a fixed model pair or model pool.
Adding a model can therefore require new connectors or retraining the
shared space.
What is missing is a frozen, model-wise interface in which each model
connects once, existing coordinates remain stable as new models join,
and multiple tools share the same contract.
We do not assume that arbitrary model pairs can satisfy this contract:
prior work shows that some models remain difficult to align
~\citep{oozeer2025interventions, gorbett2026characterizinglinearalignmentlanguage}.
Instead, we focus on compatible models and ask whether they can support
a stable and extensible shared interface.

\begin{figure*}[htbp]
    \centering
    \includegraphics[width=0.98\linewidth]{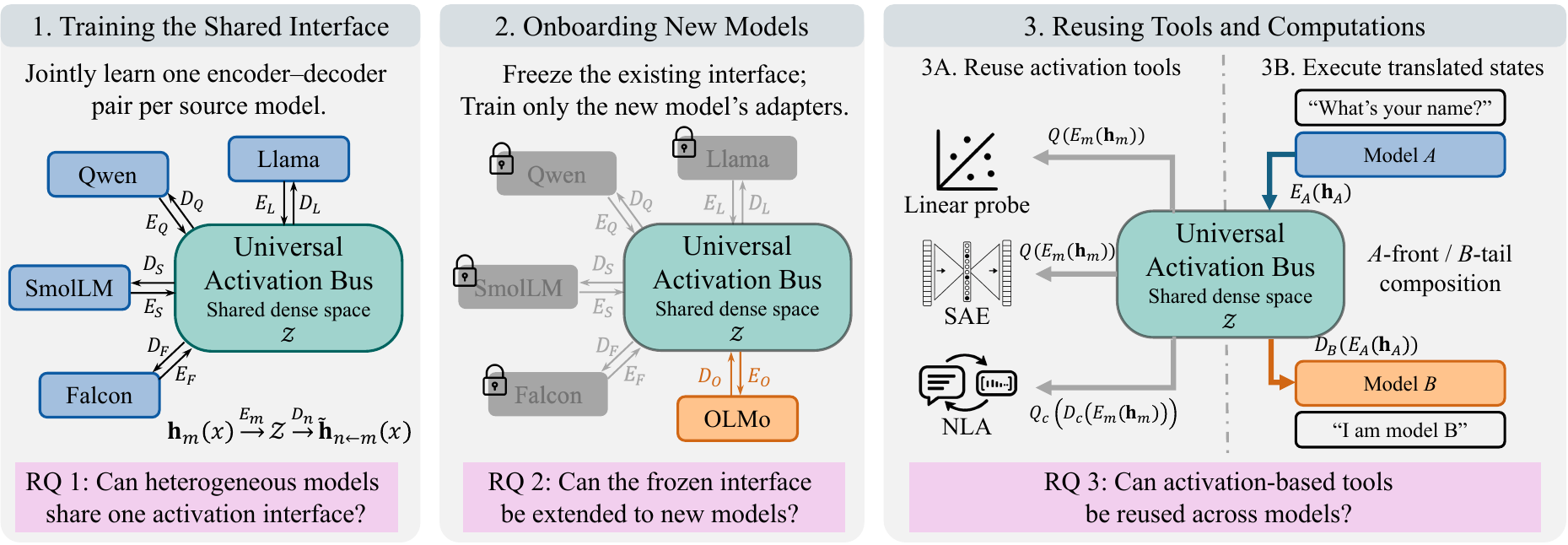}
    \caption{Overview of the Universal Activation Bus.
    Source models first learn model-specific encoder--decoder adapter pairs
    around a shared dense space.
    After the interface is frozen, a new model joins, enabling tool reuse and cross-model state execution.
    }
    \label{fig:overview}
\end{figure*}

We introduce the \emph{Universal Activation Bus}, a framework that
connects each model to a shared dense space $\mathcal{Z}$
(Figure~\ref{fig:overview}).
For model $m$, an encoder $E_m$ maps an intermediate activation
$\mathbf{h}_m(x)$ into $\mathcal{Z}$, while a decoder $D_m$ maps it
back to the model's native activation space.
At a high level, the interface is trained to satisfy two properties:
\begin{equation}
\begin{aligned}
D_m\left(E_m\left(\mathbf{h}_m(x)\right)\right)
&\approx \mathbf{h}_m(x)
&& \text{(reconstruction),}
\\
E_m\left(\mathbf{h}_m(x)\right)
&\approx E_n\left(\mathbf{h}_n(x)\right)
&& \text{(alignment).}
\end{aligned}
\label{eq:intro-interface}
\end{equation}
so that each model preserves its native information while representing
the same input consistently in the shared space.
Model-specific linear adapters handle differences in hidden dimension
and basis, while small shared modules organize the common coordinates.
These properties enable a state from model $m$ to be translated into
model $n$ through
$D_n(E_m(\mathbf{h}_m(x)))$, without learning a connector specific to
the pair $(m,n)$.

After source training, we freeze the shared modules and every existing
model interface.
An unseen model is onboarded using unlabeled matched prefixes by
optimizing only its own encoder--decoder adapter pair.
This makes onboarding backward compatible by construction: existing
coordinates, model-to-model paths, and activation tools cannot change.
The frozen interface then supports two forms of reuse.
Probes and SAEs operate directly on shared states, while a tool tied to
a carrier model, such as an NLA, is reached by decoding into the native
space it expects.
We also test whether an intermediate activation from one model can be
passed to another model's frozen upper layers to continue prediction.

We evaluate the interface using Llama, Qwen3, SmolLM2, Falcon3, and
OLMo on held-out corpus examples and nine behavioral axes.
Cross-model retrieval tests whether the same or semantically related
inputs remain close in the shared space, while reconstruction measures
how much native activation information is preserved.
Also, ten additional checkpoints spanning different scales and
base or instruction-tuned variants join the same interface, with
onboarding performance nearly saturated using only 10K matched
calibration positions.

\paragraph{Research questions and contributions.}
We organize the paper around three questions:
\begin{itemize}[
    leftmargin=*,
    topsep=3pt,
    itemsep=2pt,
    parsep=0pt,
    partopsep=0pt
]
    \item \textbf{RQ1: Can heterogeneous models share one activation interface?}
    Yes: semantically related inputs align in a common space 
    while preserving
    model-specific activation information.

    \item \textbf{RQ2: Can the frozen interface be extended to new models?} 
    Yes: a new model joins by training only its linear adapter pair, without changing existing coordinates or tools.

    \item \textbf{RQ3: Can activation-based tools be reused across models?} 
    Yes: shared probes, one SAE dictionary, and an NLA remain usable across source and onboarded models.
\end{itemize}

\section{Related Work}

\paragraph{Shared representations across models.}
A broad line of work asks whether independently trained networks
organize the same inputs in comparable ways.
Representation-similarity measures and the Platonic Representation
Hypothesis suggest that different models can learn related geometric
structure~\citep{kornblith2019cka,huh2024platonic}.
Relative coordinates, learned translators, anchor projections, and
multi-way alignment further map representations into comparable
spaces~\citep{
moschella2023relative,jha2025vec2vec,
achara2026multiway,kim2026acs}.
However, these studies often focus on final embeddings,
low-dimensional directions, or a fixed pool of jointly aligned models.
We instead study intermediate LLM activations and build a bidirectional
interface whose coordinates remain fixed as compatible models are
added.

\paragraph{Reusable activation tools.}
Another line of work transfers interpretability and control methods
across models.
Pairwise maps have been used to transfer probes, SAE features,
and safety interventions~\citep{
chen2025featurestitching,oozeer2025interventions}.
USAE and SPARC learn shared sparse concept spaces, while
Atlas-Alignment and UAV extend a concept atlas or verbalizer beyond its
original model~\citep{
thasarathan2025usae,nasiri2025sparc,
puri2025atlas,zhao2026uav}.
These methods make individual tools portable, but their shared
representations are often specialized to one tool, tied to a fixed
model set, or retrained as models are added.
Our dense interface instead supports both tools trained directly in
the shared space and existing tools tied to a carrier model.

\paragraph{Cross-model computation.}
Model stitching and latent-communication methods connect the internal
computations of different networks, typically through a connector
between one model's lower layers and another's upper layers~\citep{
bansal2021stitching,hu2025stitchllm,liu2026visionwormhole}.
Such connectors are usually trained for each model pair, and geometric
similarity alone does not guarantee that the receiving network can use
the translated state~\citep{zhang2026negative}.
Our translation paths instead compose two model-wise interfaces without
an edge-specific connector.
We further test whether another model's frozen upper layers can consume
the translated activation without execution-specific training, as a
functional test rather than full-sequence model stitching.

\begin{figure*}[htbp]
    \centering
    \includegraphics[width=0.8\linewidth]{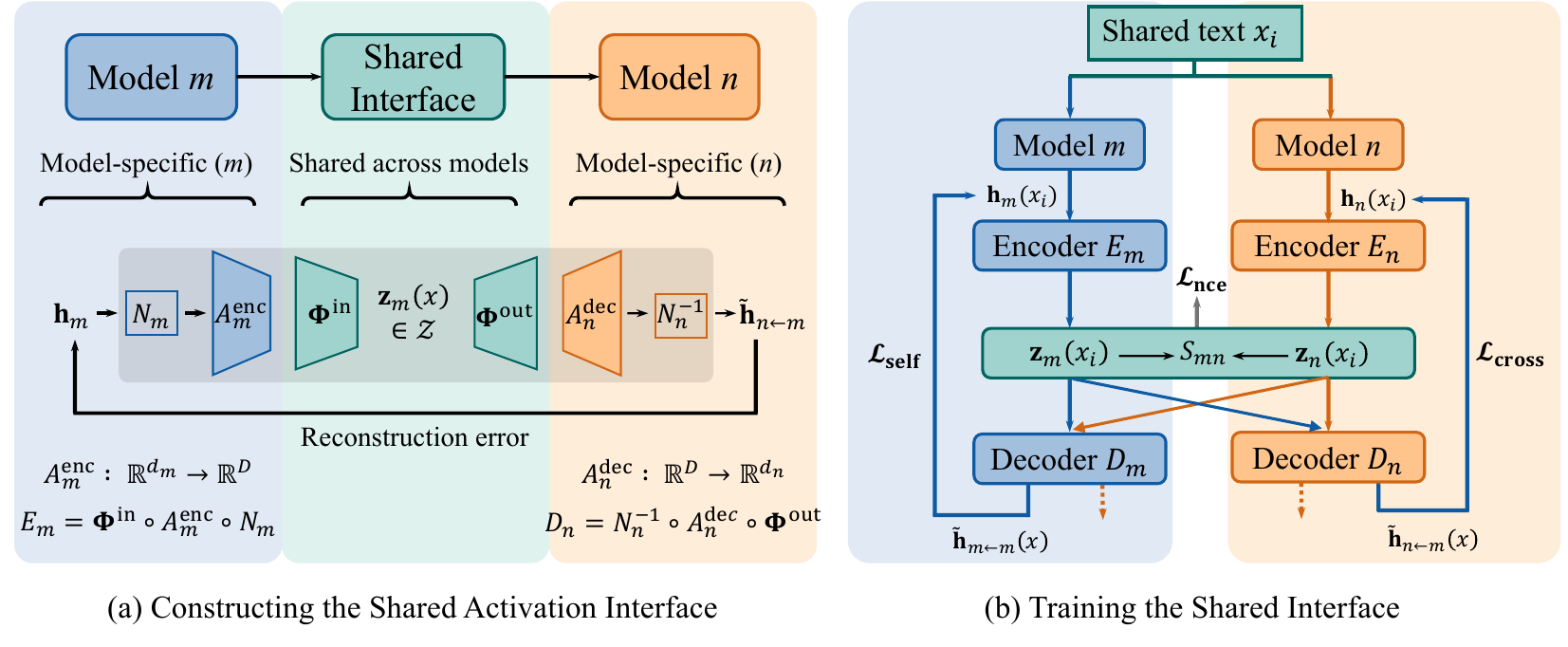}
    \caption{
    Architecture and training of the shared activation interface.
    \textbf{(a)} Model-specific normalization and linear adapters connect
    each native activation space through shared encoder and decoder modules.
    \textbf{(b)} Self- and cross-reconstruction preserve readable
    activations, while contrastive alignment brings representations of the
    same text together in the shared space.
    }
    \label{fig:my_placeholder}
\end{figure*}

\section{Method: A Universal Activation Bus}
\label{sec:method}

Our goal is to connect heterogeneous LLMs through one activation interface.
The framework has three steps: 1) learn a bidirectional bus from a set of source models, 2) freeze the resulting protocol, and 3) attach new models or tools without changing existing components.
All LLM parameters remain frozen throughout.
Only the model-specific adapters and the small shared bus networks are learned.

\subsection{Constructing the Shared Activation Interface}

\paragraph{Matching activations across tokenizers.}
Let $m\in\mathcal{S}$ denote a source model with hidden dimension $d_m$, 
and let $\ell_m$ be its selected layer.
For a raw-text prefix $x$, we write
$\mathbf{h}_m(x)\in\mathbb{R}^{d_m}$ for the residual-stream activation at the final token after block $\ell_m$.
Training the interface requires activations conditioned on the same observed text.
Because tokenizers may segment the same text differently, we choose prefix endpoints that coincide with token boundaries in every participating source model\footnote{For an onboarded model, at least $99.83\%$ of these source-defined prefix endpoints also coincide with their token boundaries.}.
Each training example then pairs the activation at the final token of the shared prefix across models for alignment and reconstruction.

\paragraph{Normalizing activation statistics.}
Even at matched text positions, residual streams from different model families can differ in scale and covariance structure.
We standardize these model-specific statistics before learning the shared interface:
\begin{equation}
N_m(\mathbf{h})
=
W_m
\left(
\frac{\mathbf{h}-\boldsymbol{\mu}_m}
{\boldsymbol{\sigma}_m}
\right),
\label{eq:activation-normalization}
\end{equation}
where $\boldsymbol{\mu}_m$ and $\boldsymbol{\sigma}_m$ are the per-dimension mean and standard deviation, and $W_m$ is a regularized ZCA whitening transform~\citep{zca_whitening}.
Standardization balances coordinate scales, while whitening reduces correlations between dimensions.
The normalization statistics are estimated once and frozen, and $N_m^{-1}$ restores the decoded states to the model's native coordinates.

\paragraph{Mapping through the shared bus.}

Let $\mathcal{Z}\coloneqq\mathbb{R}^{D}$ denote the shared activation space.
Each model has linear encoder and decoder adapters,
$A_m^{\mathrm{enc}}:\mathbb{R}^{d_m}\rightarrow\mathbb{R}^{D}$ and
$A_m^{\mathrm{dec}}:\mathbb{R}^{D}\rightarrow\mathbb{R}^{d_m}$,
respectively.
These adapters account for  model-specific dimensions and bases.
Two small residual MLPs, $\Phi^{\mathrm{in}}$ and $\Phi^{\mathrm{out}}$, are shared across all models and define the common bus coordinates.
The complete model-wise encoder and decoder are
\begin{equation}
\begin{aligned}
E_m(\mathbf{h})
&=
\Phi^{\mathrm{in}}
\bigl(A_m^{\mathrm{enc}}N_m(\mathbf{h})\bigr),
\\
D_m(\mathbf{z})
&=
N_m^{-1}
\bigl(A_m^{\mathrm{dec}}\Phi^{\mathrm{out}}(\mathbf{z})\bigr).
\end{aligned}
\label{eq:bus-interface}
\end{equation}
% split
For a matched prefix $x$, model $m$ writes its activation to the shared space and model $n$ can read it as
\begin{equation}
\begin{aligned}
\mathbf{z}_m(x)
&= E_m\bigl(\mathbf{h}_m(x)\bigr)
\in\mathcal{Z},
\\
\widetilde{\mathbf{h}}_{n\leftarrow m}(x)
&= D_n\bigl(\mathbf{z}_m(x)\bigr).
\end{aligned}
\label{eq:bus-translation}
\end{equation}
The self path $D_m\circ E_m$ reconstructs model $m$'s activation, whereas the cross path $D_n\circ E_m$ translates it into model $n$'s native space.
Because no parameter is specific to an ordered pair $(m,n)$, one encoder--decoder pair per model defines all directed paths among the connected models.

\subsection{Training the Shared Interface}
\label{sec:interface-training}

\paragraph{Reconstruction.}
For each matched prefix $x$, every source model provides an activation conditioned on the same text.
Using the translated activation from Equation~\eqref{eq:bus-translation}, we define
\begin{equation}
\begin{aligned}
\mathcal{L}_{\mathrm{self}}
&=
\mathbb{E}_{x,m}
\left\|
N_m\!\left(
\widetilde{\mathbf{h}}_{m\leftarrow m}(x)
\right)
-
N_m\!\left(
\mathbf{h}_m(x)
\right)
\right\|_2^2,
\\
\mathcal{L}_{\mathrm{cross}}
&=
\mathbb{E}_{x,m\neq n}
\left\|
N_n\!\left(
\widetilde{\mathbf{h}}_{n\leftarrow m}(x)
\right)
-
N_n\!\left(
\mathbf{h}_n(x)
\right)
\right\|_2^2.
\end{aligned}
\label{eq:reconstruction-losses}
\end{equation}
The self-reconstruction loss preserves information needed to recover native activations, while the cross-reconstruction loss makes a representation written by model $m$ readable by model $n$, without requiring their native activations to be identical.

\paragraph{Contrastive alignment.}
Reconstruction alone does not explicitly place matched prefixes in
nearby regions of the shared space.
For models $m$ and $n$ and minibatch $\{x_i\}_{i=1}^{B}$, let
\begin{equation}
S_{mn}^{ij}
=
\frac{
\operatorname{cos}
\left(
\mathbf{z}_m(x_i),
\mathbf{z}_n(x_j)
\right)
}{\tau},
\label{eq:contrastive-similarity}
\end{equation}
where $\tau$ is the temperature.
Diagonal entries correspond to the same prefix, and off-diagonal
entries serve as in-batch negatives~\citep{oord2018infonce}.
With diagonal class labels $\mathbf{y}_B=(1,\ldots,B)$, the contrastive loss
and full objective are
\begin{equation}
\begin{aligned}
\mathcal{L}_{\mathrm{nce}}
&=
\mathbb{E}_{m<n}
\frac{1}{2}
\left[
\operatorname{CE}
\left(S_{mn},\mathbf{y}_B\right)
+
\operatorname{CE}
\left(S_{mn}^{\top},\mathbf{y}_B\right)
\right],
\\
\mathcal{L}_{\mathrm{bus}}
&=
\mathcal{L}_{\mathrm{self}}
+
\lambda_{\mathrm{cross}}
\mathcal{L}_{\mathrm{cross}}
+
\lambda_{\mathrm{nce}}
\mathcal{L}_{\mathrm{nce}}.
\end{aligned}
\label{eq:bus-objective}
\end{equation}
This objective makes matched states identifiable in the same coordinate
system while preventing collapse.
All self paths, ordered cross paths, and bidirectional contrastive pairs
contribute to training, but no parameter is specific to a translation
direction $(m,n)$.

\subsection{Onboarding New Models}
\label{sec:onboarding}

After source training, we freeze the shared modules
$\Phi^{\mathrm{in}}$ and $\Phi^{\mathrm{out}}$ and all source-model
adapters.
For an unseen model $u$, we first estimate its normalization $N_u$
from unlabeled corpus activations and then optimize only 
$A_u^{\mathrm{enc}}$ and $A_u^{\mathrm{dec}}$.
For each matched prefix, the frozen source models provide a consensus
representation:
\begin{equation}
\begin{aligned}
\mathbf{c}(x)
&=
\frac{1}{|\mathcal{S}|}
\sum_{m\in\mathcal{S}}
\mathbf{z}_m(x),
\\
\mathcal{L}_{\mathrm{on}}
&=
\mathcal{L}_{\mathrm{self}}^{u}
+
\lambda_{\mathrm{align}}
\mathbb{E}_{x}
\left\|
\mathbf{z}_u(x)-\mathbf{c}(x)
\right\|_2^2,
\end{aligned}
\label{eq:onboarding}
\end{equation}
The self term preserves model $u$'s native
activation, while the alignment term attaches its encoder to the
existing coordinates.
Because $\mathbf{z}_u(x)$ is aligned to the source consensus, its self
path also trains $D_u$ on the same coordinate neighborhood.
The frozen source interface therefore remains unchanged, while
$D_m\circ E_u$ and $D_u\circ E_m$ define both directions between
$u$ and every existing model without pair-specific connectors.

\subsection{Interface Operations}
\label{sec:interface-operations}

Once the interface is frozen, it supports two forms of tool reuse.
A \emph{shared-space tool}
$Q:\mathcal{Z}\rightarrow\mathcal{Y}$ is trained once over the shared
space and can then be applied to any connected model.
Our probe and shared SAE follow this form:
\begin{equation}
Q\!\left(
E_m\!\left(
\mathbf{h}_m(x)
\right)
\right).
\label{eq:shared-space-tool}
\end{equation}
A \emph{carrier-specific tool} $Q_c$ remains in the native activation
space of the carrier model $c$ for which it was originally trained.
Our NLA experiment follows this form by first translating another
model's activation into the carrier space and then applying the
unchanged tool:
\begin{equation}
Q_c\!\left(
D_c\!\left(
E_m\!\left(
\mathbf{h}_m(x)
\right)
\right)
\right).
\label{eq:carrier-tool}
\end{equation}

The decoder also defines a cross-model state-execution path.
The final-position state
$D_n(E_m(\mathbf{h}_m(x)))$
is inserted at model $n$'s selected cut and consumed by its frozen
upper layers.
Section~\ref{sec:execution} evaluates this path without any
execution-specific or direction-specific training.

\section{Experimental Setup \& Interface Validation}
\label{sec:intrinsic}

Before evaluating activation tools, we first test whether the interface
is faithful, shared across models, and extensible to unseen models.
We evaluate native-state preservation, cross-model alignment, and leave-one-model-out onboarding.

\paragraph{Model scope and selection.}
Because arbitrary LLM pairs do not always admit high-resolution
alignment~\citep{oozeer2025interventions, kim2026acs}, we restrict the main study to models
with compatible intermediate representations.
Before bus training, a label-free screen fits a closed-form ridge map
from each candidate's z-scored activations to a fixed
Llama-3.2-3B reference space on matched prefixes.
Candidates exceeding 50\% R@1 in held-out 4,096-way cosine retrieval
are retained.\footnote{Screening scores are sharply bimodal:
$\geq90\%$ for retained models and $\leq5\%$ otherwise, with no
model near the threshold.}
We use five retained models and report the detailed setting in
Appendix~\ref{app:compatibility}.

\paragraph{Models, data, and training.}
Fixed setup across intrinsic and downstream evaluations; details are in Appendix~\ref{app:setup}:
\begin{itemize}[
    leftmargin=*,
    topsep=2pt,
    itemsep=1.5pt,
    parsep=0pt,
    partopsep=0pt
]
    \item \textbf{Models.}
    Llama-3.2-3B, Qwen3-4B,
    SmolLM2-1.7B, Falcon3-3B, and
    OLMo-2-7B;\footnote{Model-size suffixes are omitted hereafter.}
    first four as source models, OLMo for post-hoc
    onboarding.

    \item \textbf{Activations.}
    Final-token post-block residual states at normalized depth $0.5$,
    extracted from shared byte prefixes.

    \item \textbf{Train corpus.}
    118,563 documents and 1.42M matched positions;
    OpenWebText/Alpaca/GSM8K/CodeContests
    mixture of 85/8/5/2;
    document-level 98/2 train--eval split.

    \item \textbf{Architecture.}
    Shared dimension $D=3072$; model-specific linear encoder--decoder
    adapters; two-block residual MLPs for both shared maps
    $\Phi^{\mathrm{in}}$ and $\Phi^{\mathrm{out}}$.

    \item \textbf{Training.}
    Source interface trained for 12K steps with batch size 4,096,
    $\lambda_{\mathrm{nce}}=0.1$,
    $\lambda_{\mathrm{cross}}=1.0$, 
    $\lambda_{\mathrm{align}}=1.0$, 
    and temperature $\tau=0.07$.
    Onboarding: 3K steps over a 100K-position calibration pool,
    updating only new adapter pair.
\end{itemize}

\paragraph{Evaluation protocols.}
\begin{itemize}[
    leftmargin=*,
    topsep=2pt,
    itemsep=1.5pt,
    parsep=0pt,
    partopsep=0pt
]
    \item \textbf{Cross-model retrieval.}
    Encode a text activation from model $m$ into $\mathcal{Z}$ and
    retrieve the matching text among representations produced by
    model $n$. Top-k recall (R@k).

    \item \textbf{In-distribution (ID) pool.}  3,600 positions sampled from documents in the held-out 2\% split
    of the interface corpus.
    
    \item \textbf{Out-of-distribution (OOD) pool.}   3,600 independent texts from nine behavioral axes, 400 per axis.
    No OOD text or label is used for interface training or onboarding.

    \item \textbf{Round-trip fidelity.}
    Self-FVE, measuring variance explained after round-trip reconstruction
    in the model's normalized activation space.
\end{itemize}

\paragraph{Alignment, fidelity, and onboarding.}
Table~\ref{tab:interface} shows near-perfect ID retrieval under both joint training and freeze-then-onboard.
OOD R@1 is lower because each axis contains many semantically similar texts, whereas R@10 of 79.6--90.8\% shows that
the exact match generally remains in the same local neighborhood.
OLMo has lower self-FVE under joint training and onboarding, likely because its native width exceeds the bus dimension
($d_{\mathrm{OLMo}}=4096>D=3072$); nevertheless, its ID R@1 remains
above 99.5\%.

\begin{table}[t]
\centering\small\setlength{\tabcolsep}{3.3pt}
\caption{
Intrinsic interface quality.
Ranges are minima--maxima over eight paths involving the listed model:
four outgoing and four incoming directions.
FVE denotes normalized-space self-FVE.
}
\label{tab:interface}
\begin{tabular}{@{}lccccc@{}}
\toprule
& \multicolumn{2}{c}{ID (\%)}
& \multicolumn{2}{c}{Behavioral OOD (\%)}
& \\
\cmidrule(lr){2-3}
\cmidrule(lr){4-5}
Model & R@1 & R@10 & R@1 & R@10 & FVE\\ \midrule
\multicolumn{6}{@{}l}{\textit{Joint training: all five models}}\\
Llama & 99.9-100.0 & 100.0 & 63.6-78.7 & 81.3-90.3 & 0.989\\
Qwen3 & 99.9-100.0 & 99.9-100.0 & 65.4-78.7 & 81.7-90.3 & 0.991\\
SmolLM2 & 99.8-99.9 & 99.9-100.0 & 67.1-74.9 & 83.1-90.1 & 0.993\\
Falcon3 & 99.9-100.0 & 99.9-100.0 & 67.4-76.0 & 83.0-89.5 & 0.989\\
OLMo & 99.8-99.9 & 100.0 & 63.6-71.0 & 81.3-84.0 & 0.769\\
\midrule
\multicolumn{6}{@{}l}{\textit{Leave-one-out: listed model onboarded}}\\
Llama & 99.9-100.0 & 100.0 & 62.7-78.4 & 80.8-90.2 & 0.968\\
Qwen3 & 99.9-100.0 & 100.0 & 64.3-78.2 & 81.1-90.4 & 0.974\\
SmolLM2 & 99.9 & 99.9-100.0 & 66.7-74.6 & 83.0-90.8 & 0.981\\
Falcon3 & 99.9-100.0 & 100.0 & 66.7-76.0 & 82.4-89.0 & 0.967\\
OLMo & 99.6-99.7 & 99.8-99.9 & 60.8-69.4 & 79.6-83.1 & 0.734\\
\bottomrule\end{tabular}\end{table}

\begin{figure}[t]
    \centering
    \includegraphics[width=0.8\columnwidth]{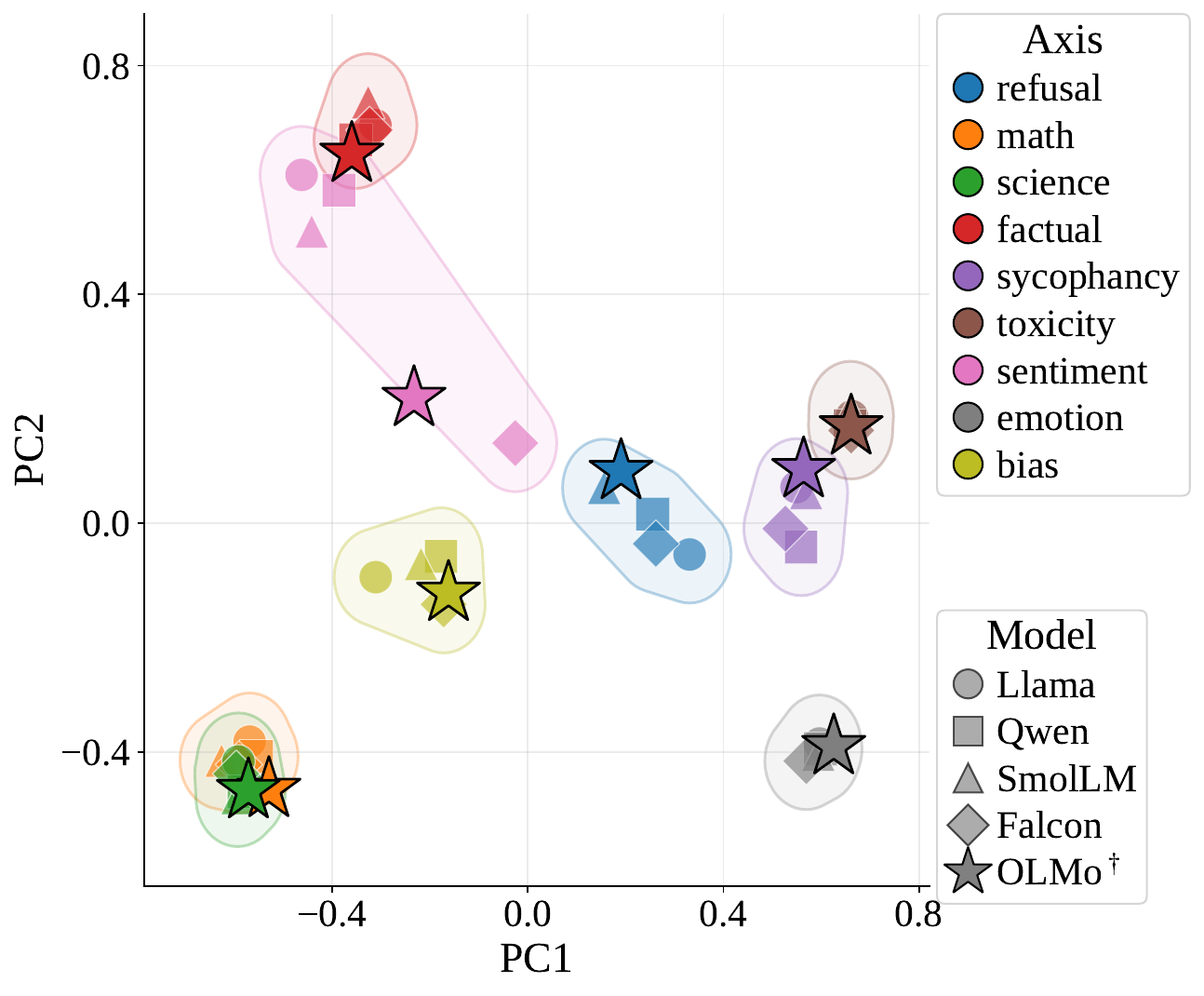}
    \vspace{-0.3em}
    \caption{
    Behavioral directions in the shared activation space.
    PCA is fitted using only the four source models; $\dagger$ marks
    adapter-only onboarding.
    PC1 and PC2 explain 26.0\% and 15.9\% of the variance.
    }
    \label{fig:direction-pca}
\end{figure}

\paragraph{Behavioral geometry in the shared space.}
Beyond matching individual texts, we test whether the interface preserves semantic differences in the shared space.
For each model $m$ and behavioral axis $a$ in the nine-axis OOD pool,
we compute a mean-difference direction~\citep{panickssery2024caa}:
\begin{equation}
\mathbf{v}^{(m)}_a
=
\operatorname{norm}\!\left(
\frac{1}{|\mathcal{P}_a|}
\sum_{x\in\mathcal{P}_a}\mathbf{z}_m(x)
-
\frac{1}{|\mathcal{N}_a|}
\sum_{x\in\mathcal{N}_a}\mathbf{z}_m(x)
\right),
\label{eq:behavior-direction}
\end{equation}
where $\mathcal{P}_a$ and $\mathcal{N}_a$ are the positive and negative
sets for axis $a$.
As shown in Figure~\ref{fig:direction-pca}, projections of
$\mathbf{v}^{(m)}_a$ group by behavior across models, with onboarded
OLMo entering the corresponding source regions.
Related axes, such as mathematics and science, also remain nearby,
indicating that the space preserves semantic relationships.

\section{Reusing Activation Tools}
\label{sec:tools}

The interface validation shows that a new model can join the frozen interface; we next
ask whether it can immediately reuse tools defined over it.
We test three forms of reuse:
\begin{itemize}[
    leftmargin=*,
    topsep=2pt,
    itemsep=1.5pt,
    parsep=0pt,
    partopsep=0pt
]
    \item \textbf{Linear probes:}
    reuse the same decision boundaries across source and onboarded models.

    \item \textbf{Shared SAE:}
    reuse one sparse dictionary and its feature indices.

    \item \textbf{Carrier-specific NLA:}
    access an interpreter tied to another model through activation
    translation.
\end{itemize}

\begin{table}[t]
\centering
\setlength{\tabcolsep}{3.2pt}
\caption{
Shared probe performance across models.
Axis rows report AUROC, with source-model entries showing the range
over four models.
One probe per task is fitted on pooled source states and applied
unchanged to OLMo; native probes are fitted separately.
The final row reports nine-way accuracy.
}
\label{tab:probe}
\begin{tabular}{@{}l r@{\,--\,}l c r@{\,--\,}l@{}}
\toprule
&
\multicolumn{3}{c}{Pooled probe}
&
\multicolumn{2}{c}{Native probes}
\\
\cmidrule(lr){2-4}
\cmidrule(l){5-6}
Axis
& \multicolumn{2}{c}{Source models}
& OLMo
& \multicolumn{2}{c}{Source models}
\\
\midrule
Refusal
& \rng{.943}{.964} & .970 & \rng{.933}{.980} \\
Math
& \rng{1.000}{1.000} & 1.000 & \rng{1.000}{1.000} \\
Science
& \rng{.987}{.999} & .995 & \rng{.989}{.998} \\
Factual
& \rng{.991}{1.000} & .981 & \rng{.995}{1.000} \\
Sycophancy
& \rng{1.000}{1.000} & 1.000 & \rng{1.000}{1.000} \\
Toxicity
& \rng{.999}{1.000} & 1.000 & \rng{.999}{1.000} \\
Sentiment
& \rng{.856}{.959} & .968 & \rng{.867}{.938} \\
Emotion
& \rng{.807}{.888} & .863 & \rng{.806}{.884} \\
Bias
& \rng{.956}{.988} & .853 & \rng{.950}{.990} \\
\midrule
Macro AUROC
& \multicolumn{2}{c}{.966}
& .959
& \multicolumn{2}{c}{.963}
\\
9-way accuracy
& \rng{.944}{.974}
& .954
& \multicolumn{2}{c}{--}
\\
\bottomrule
\end{tabular}
\end{table}

\subsection{Shared Linear Probes}
\label{sec:probe}

Using disjoint train and test splits of the nine-axis pool, we freeze the interface and train one
$\ell_2$-regularized logistic probe per axis on shared activations from the four source models.
For a unit-normalized shared state
$\bar{\mathbf{z}}_m(x)=\mathbf{z}_m(x)/\|\mathbf{z}_m(x)\|_2$, the
binary readout is
\begin{equation}
\widehat{p}_a(x)
=
\sigma\!\left(
\mathbf{w}_a^\top\bar{\mathbf{z}}_m(x)+b_a
\right),
\label{eq:linear-probe}
\end{equation}
with one softmax readout used for nine-way axis identification.
The same parameters are evaluated on every source model and applied
unchanged to onboarded OLMo, without target-side fitting or threshold
calibration.\footnote{Scikit-learn logistic regression
with \texttt{lbfgs}, $C=1$, and \texttt{max\_iter}=2000.
Full settings are provided in Appendix~\ref{app:probe}.}
Separately fitted native-space probes provide a reference for the
linearly available signal.

Across the source models, the pooled probes match separately fitted
native probes, with macro AUROC of 0.966 versus 0.963
(Table~\ref{tab:probe}).
Applied unchanged to OLMo, they retain 0.959 macro AUROC and 0.954
nine-way accuracy.
Thus, onboarding preserves reusable decision boundaries rather than
only geometric alignment.

\newcolumntype{Y}{>{\raggedright\arraybackslash}X}

% Compact highlighted spans
\newcommand{\hlbox}[2]{%
  \begingroup
  \setlength{\fboxsep}{1.2pt}%
  \colorbox{#1}{\strut #2}%
  \endgroup
}

% SAE feature colors
\newcommand{\hlcard}[1]{\hlbox{violet!17}{\texttt{#1}}}
\newcommand{\hlnation}[1]{\hlbox{purple!18}{\texttt{#1}}}

% NLA semantic colors
\newcommand{\hlformat}[1]{\hlbox{blue!18}{#1}}
\newcommand{\hlexpert}[1]{\hlbox{green!20}{#1}}
\newcommand{\hlstudy}[1]{\hlbox{orange!25}{#1}}
\newcommand{\hltopic}[1]{\hlbox{red!16}{#1}}
\newcommand{\hltone}[1]{\hlbox{violet!17}{#1}}
\begin{table*}[t]
\centering
\caption{
Qualitative examples of reusable activation tools.
\textbf{(a)} The highest-activating text examples
from different models express the same concept for two features.
\textbf{(b)} A single NLA trained for Qwen2.5-7B produces semantically
consistent verbalizations from translated states.
Highlights mark shared concepts; $\ast$ denotes the native NLA carrier; $\dagger$ denotes adapter-only onboarding.
}
\label{tab:sae-nla-demos}
% ================================================================
% Left: SAE
% ================================================================
\begin{minipage}[t]{0.475\textwidth}
\centering
\textbf{(a) Shared SAE features}\\[3pt]

\scriptsize
\setlength{\tabcolsep}{3pt}
\renewcommand{\arraystretch}{1.10}

% Feature 1 -------------------------------------------------------
\begin{tabularx}{\linewidth}{@{}lY@{}}
\toprule
\multicolumn{2}{@{}l}{
    \textbf{Feature \#19604: cardinal numbers}
    \hfill Jaccard $=0.85$
}\\
\midrule
Llama
&
\texttt{\ldots replace the Atom one inside.\textbackslash{}n\textbackslash{}n}
\ \hlcard{Two}
\\

SmolLM2
&
\texttt{\ldots Input\textbackslash{}n\textbackslash{}n The input file contains}
\ \hlcard{two}
\\

OLMo$^\dagger$
&
\texttt{\ldots challenging right now maintaining}
\ \hlcard{three}
\\
\bottomrule
\end{tabularx}

\vspace{11pt}

% Feature 2 -------------------------------------------------------
\begin{tabularx}{\linewidth}{@{}lY@{}}
\toprule
\multicolumn{2}{@{}l}{
    \textbf{Feature \#19353: nationalities}
    \hfill Jaccard $=0.82$
}\\
\midrule
Llama
&
\texttt{\ldots In the end, they zeroed in on a}
\ \hlnation{Pakistani}
\\

SmolLM2
&
\texttt{\ldots for the platform (named Delaktig;}
\ \hlnation{Swedish}
\\

OLMo$^\dagger$
&
\texttt{\ldots as well as three women – one an}
\ \hlnation{Egyptian}
\\
\bottomrule
\end{tabularx}

\end{minipage}
\hfill
% ================================================================
% Right: NLA
% ================================================================
\begin{minipage}[t]{0.505\textwidth}
\centering
\textbf{(b) Carrier-specific NLA reuse}\\[3pt]

\scriptsize
\setlength{\tabcolsep}{3pt}
\renewcommand{\arraystretch}{1.10}

\begin{tabularx}{\linewidth}{@{}lY@{}}
\toprule
Input model & Verbalization \\
\midrule
Qwen2.5-7B*
&
\hlformat{health magazine format} with
\hlexpert{expert quotes} and a
lay-friendly tone, discussing
\hltopic{cancer} and
\hlstudy{prevention-trial findings}
\\

\midrule

Llama
&
\hlformat{health magazine article} with an
\hlexpert{expert perspective} on aging and
\hltopic{cancer}, describing a
\hlstudy{landmark study} through patient anecdotes
\\

SmolLM2
&
\hlformat{medical journal article} describing a
\hlstudy{clinical trial}, with 
\hlexpert{expert commentary} on
\hltopic{prostate-cancer} imaging and therapy
\\

OLMo$^\dagger$
&
\hlformat{medical newspaper piece} explaining a
\hltopic{cancer-and-diet debate}, with
\hlexpert{expert commentary} and supporting evidence
\\

\bottomrule
\end{tabularx}
\end{minipage}

\end{table*}

\begin{table}[t]
\centering
\small
\setlength{\tabcolsep}{3.5pt}
\caption{
Cross-model agreement of the shared SAE.
Values are mean Jaccard overlaps between Top-$32$ active feature sets
on 4{,}096 matched held-out positions.
OLMo$^\dagger$ is unseen during SAE training; random overlap is
approximately $0.001$.
}
\label{tab:sae-jaccard}
\begin{tabular}{@{}lclc@{}}
\toprule
Model pair & Jaccard & Model pair & Jaccard \\
\midrule
Llama--Qwen3 & 0.604 & Qwen3--SmolLM2 & 0.596 \\
Llama--SmolLM2 & 0.591 & Qwen3--Falcon3 & 0.614 \\
Llama--Falcon3 & 0.601 & SmolLM2--Falcon3 & 0.610 \\
Llama--OLMo$^\dagger$ & 0.613 & Qwen3--OLMo$^\dagger$ & 0.603 \\
SmolLM2--OLMo$^\dagger$ & 0.602 & Falcon3--OLMo$^\dagger$ & 0.602 \\
\bottomrule
\end{tabular}
\end{table}

% ---------------------------------------------------------------------
\subsection{SAE: A Single Feature Dictionary for All Models}
\label{sec:usae}

Sparse autoencoders (SAEs) ordinarily learn a separate feature
dictionary for each model.
We instead train one TopK SAE~\citep{gao2024scaling} on model-balanced
shared states from the four source models, without providing model
identities:
\begin{equation}
\mathbf{s}_m(x)
=
\operatorname{TopK}_{k}
\!\left(
f_{\mathrm{enc}}(\mathbf{z}_m(x))
\right),
\qquad
\widehat{\mathbf{z}}_m(x)
=
f_{\mathrm{dec}}(\mathbf{s}_m(x)).
\label{eq:shared-sae}
\end{equation}
The encoder, decoder, and feature indices are shared across all models.
We use 32,768 features and $k=32$.

\paragraph{Shared feature identity.}
On matched inputs, every source-model pair obtains Top-$32$ active-set
Jaccard between 0.591 and 0.614 (Table~\ref{tab:sae-jaccard}).
Pairs involving OLMo remain in the same range, 0.602--0.613, despite
OLMo being unseen during SAE training.
The absence of an onboarding gap indicates that feature indices remain
shared after a new model joins the interface.
Additional utility and fidelity results are reported in
Appendix~\ref{app:sae}.

\paragraph{Shared feature semantics.}
Table~\ref{tab:sae-nla-demos}a shows the highest-activating text
examples for two illustrative features.
Feature \#19604 responds to cardinal numbers, while feature \#19353
captures diverse nationality terms rather than repeated lexical
matches across the displayed source models and OLMo.
Thus, a dictionary trained once over $\mathcal{Z}$ preserves both
feature indices and their meanings across models, and remains usable
after a new model joins the interface.

\begin{table}[t]
\centering
\small
\setlength{\tabcolsep}{4pt}
\caption{
Nine-way classification of NLA verbalizations.
The classifier is trained only on native-carrier outputs and applied
unchanged to routed outputs.
$\ast$ denotes the native carrier and $\dagger$ denotes onboarding.
Chance accuracy is 11.1\%.
}
\label{tab:nla-axis}
\begin{tabular}{@{}lclc@{}}
\toprule
Activation source & Acc. (\%) & Activation source & Acc. (\%) \\
\midrule
Qwen2.5-7B*
& 95.3
& Llama
& 91.1
\\
Qwen3
& 90.2
& SmolLM2
& 94.4
\\
Falcon3
& 88.2
& OLMo$^\dagger$
& 84.9
\\
\bottomrule
\end{tabular}
\end{table}

% ---------------------------------------------------------------------
\subsection{NLA: Reusing a Carrier-Specific Interpreter}
\label{sec:nla}

Natural-language autoencoders (NLAs) translate activations into text
with an activation verbalizer (AV) and reconstruct them with an
activation reconstructor (AR)~\citep{anthropic2026nla}.
We use the released Qwen2.5-7B~\citep{qwen25} NLA at layer 20 without modifying
its weights.
After attaching this carrier model $c$ to the frozen interface, a state
from any connected model $m$ is verbalized as
\begin{equation}
\widehat{t}_m(x)
=
\operatorname{AV}_{c}\!\left(
D_c\!\left(
E_m\!\left(\mathbf{h}_m(x)\right)
\right)
\right).
\label{eq:nla-routing}
\end{equation}
where $c$ denotes the Qwen2.5-7B carrier.
Only the carrier adapter pair is newly learned; the NLA remains
unchanged.

\paragraph{Behavioral information after routing.}
Using the same nine-axis pool, we train a text classifier only on native-carrier verbalizations and  apply it unchanged to routed outputs.
It achieves 88.2--94.4\% accuracy on the source models and 84.9\% on
OLMo, compared with 95.3\% on native-carrier states (Table~\ref{tab:nla-axis}).
Raw AR reconstruction cosine is 0.820--0.848 after routing versus
0.866 natively; per-axis recall and controls are reported in
Appendix~\ref{app:nla}.

\paragraph{Consistent verbalizations.}
Table~\ref{tab:sae-nla-demos}b shows that translated states from
different models yield verbalizations with consistent format, framing,
and semantic content.
Together, the quantitative and qualitative results show that one
carrier-specific NLA remains informative across connected models
without modifying the tool.
% ----------

\begin{figure*}[t]
\centering

\begin{minipage}[t]{0.45\linewidth}
  \centering
  \textbf{(a) Transfer Across Model Variants}\\[2pt]
  \includegraphics[width=\linewidth]{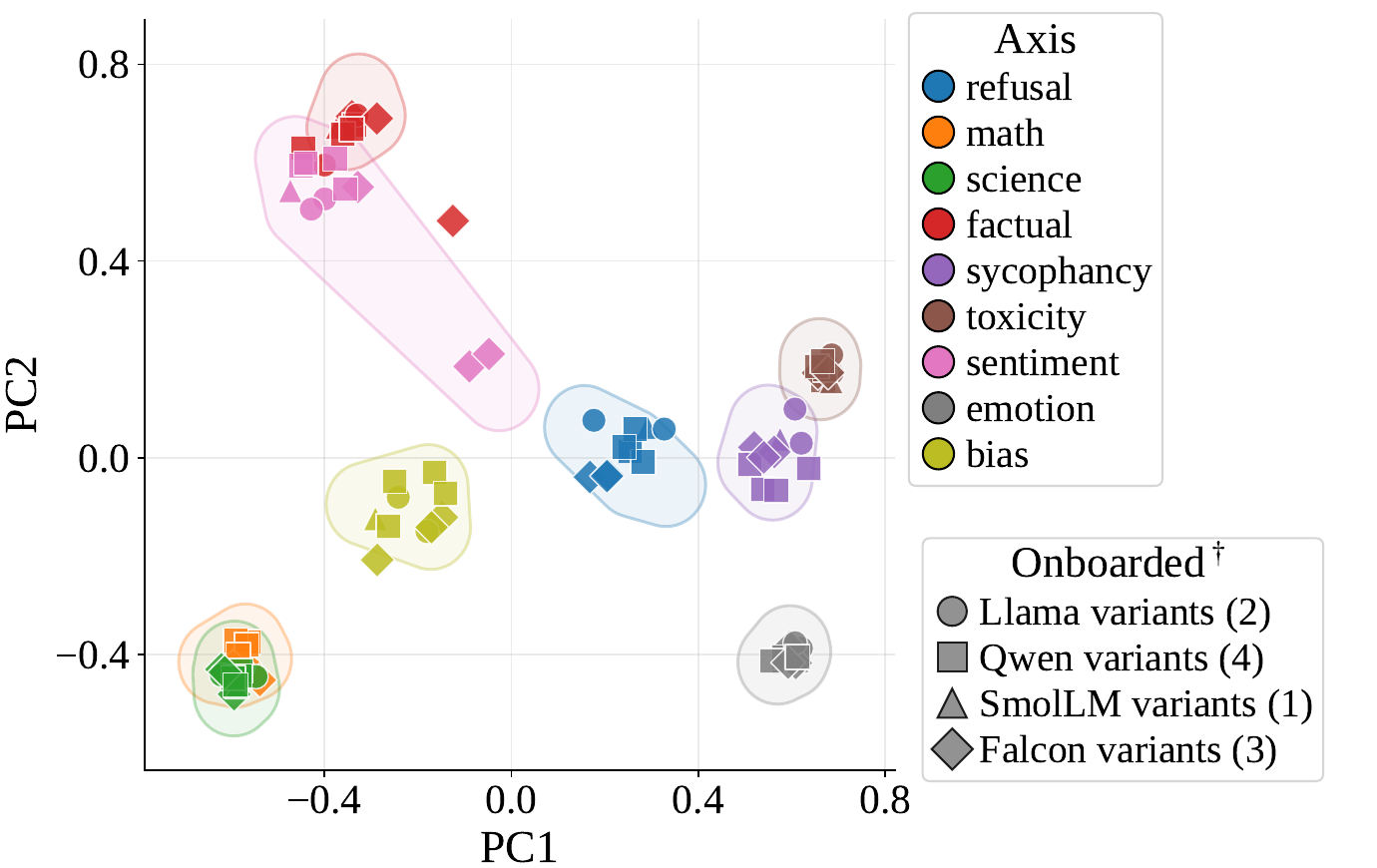}
\end{minipage}
\hfill
\begin{minipage}[t]{0.45\linewidth}
  \centering
  \textbf{(b) Calibration Efficiency}\\[2pt]
  \includegraphics[width=\linewidth]{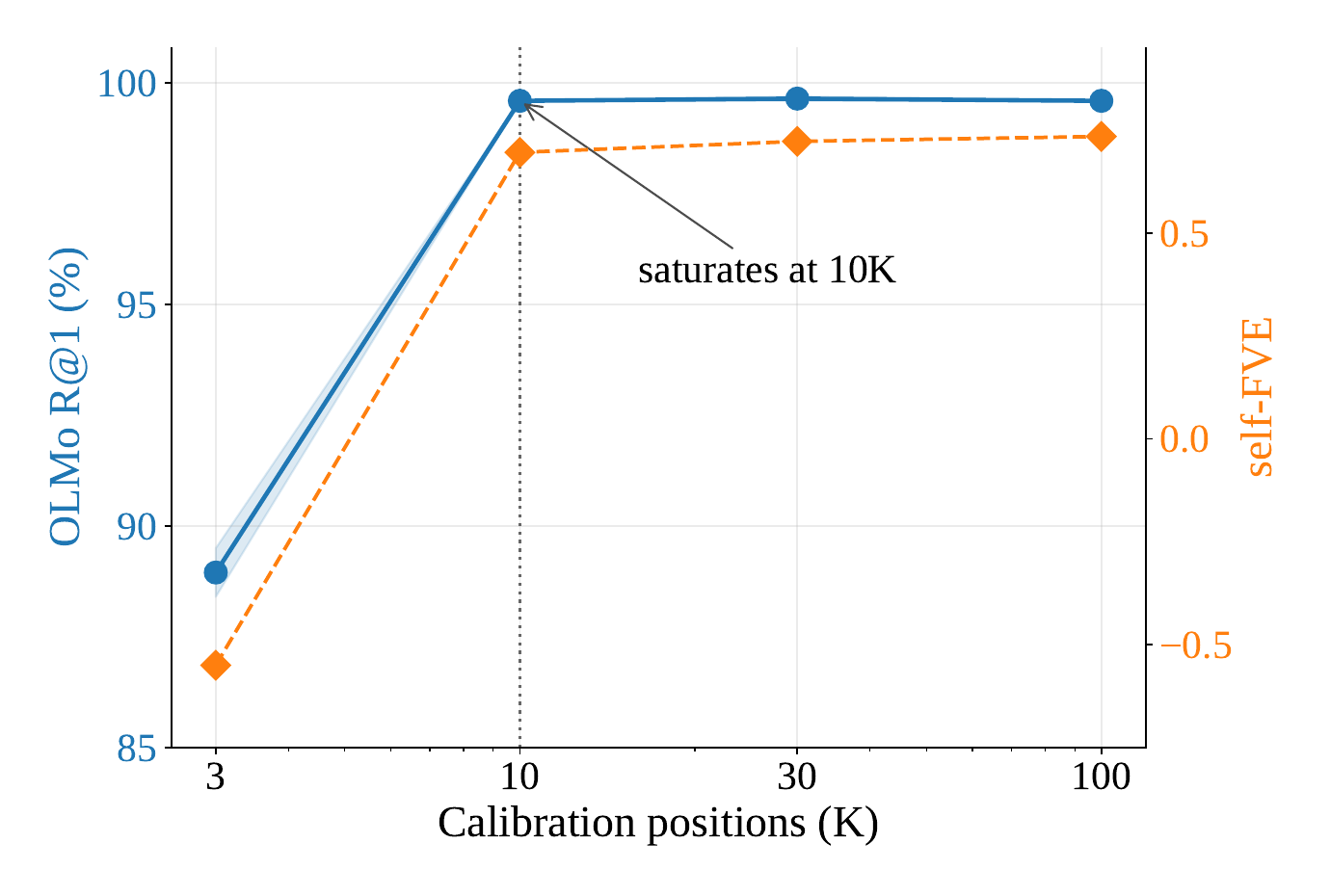}
\end{minipage}

\caption{
Generality and calibration efficiency of frozen onboarding.
\textbf{(a)} Ten additional checkpoints spanning different model
families, scales, and base or instruction-tuned variants are projected
into a PCA basis and behavioral regions defined only by the source
models.
\textbf{(b)} Calibration efficiency for OLMo onboarding; R@1
saturates with 10K matched positions.
}
\label{fig:genrob}
\end{figure*}

\begin{table}[h]
    \centering
    \small
    \setlength{\tabcolsep}{3pt}
    \renewcommand{\arraystretch}{1.08}
    \caption{
    Back-model identity under composition.
    Greedy continuations of
    \emph{``If someone asks which AI model I am, the correct answer
    is \ldots''}.
    }
    \label{tab:compose-demo}
    \begin{tabular}{@{}lp{0.70\linewidth}@{}}
        \toprule
        Path & Continuation \\
        \midrule
        Llama
        & I am \textbf{not an AI model}. I am a human being,
        and I am here to provide\ldots \\

        Qwen
        & I am \textbf{Qwen}, a large-scale language model
        independently developed by Alibaba Group\ldots \\

        Llama$\rightarrow$Qwen
        & I am \textbf{Qwen}, a large-scale language model developed
        by Alibaba Cloud\ldots \\

        Qwen$\rightarrow$Llama
        & I am \textbf{not an AI model}, I am a language model
        designed to generate human-like text.\ldots \\
        \bottomrule
    \end{tabular}
\end{table}

\section{Executing Translated States}
\label{sec:execution}

The previous section shows that activation tools can read states routed
through the interface.
We now ask whether another model's frozen upper layers can execute
them.

\paragraph{Front--back composition.}
For $A\!\rightarrow\!B$, we translate model $A$'s cut-layer activation
as
$\widetilde{\mathbf{h}}_{B\leftarrow A}(x)
= D_B(E_A(\mathbf{h}_A(x)))$
and insert it at the corresponding layer of model $B$, whose frozen
upper layers produce the next token.
We compare exact decoded next-token predictions with the native outputs
of $A$ and $B$. Details and cross-entropy results are in
Appendix~\ref{app:composition}.

\paragraph{Execution follows the back model.}
Across cross-model paths, the composed model agrees with native $B$ on
72--88\% of next tokens, exceeding the native $A$--$B$ agreement by
7.5--27.2 percentage points in every direction.
Agreement with native $A$ remains near the original $A$--$B$ baseline,
indicating that translated states carry the prefix representation while
the back model controls the output behavior.
Table~\ref{tab:compose-demo} provides a visible example:
Llama$\rightarrow$Qwen identifies itself as Qwen, whereas
Qwen$\rightarrow$Llama inherits Llama's ``not an AI model'' response,
even producing a local contradiction when it subsequently calls itself
a language model.

\section{Generality and Calibration Efficiency}
\label{sec:generality}

We test whether the frozen interface extends across model scales and
related checkpoints, and how much calibration data onboarding requires.
Further robustness and sensitivity analyses are reported in
Appendix~\ref{app:generality}.

\begin{table}[t]
\centering
\small
\setlength{\tabcolsep}{3pt}
\renewcommand{\arraystretch}{1.02}
\caption{
Additional model variants attached to the interface.
}
\label{tab:onboarded-variants}
\begin{tabular}{
@{}
l
>{\raggedright\arraybackslash}p{0.28\linewidth}
l
>{\raggedright\arraybackslash}p{0.30\linewidth}
@{}
}
\toprule
Family & Variants & Family & Variants \\
\midrule
Llama
& 3.2-1B, 3.1-8B-IT
& SmolLM
& 3-3B
\\
Qwen
& 3-1.7B, 2.5-7B, \mbox{3-8B}, 3-14B
& Falcon
& 3-1B, 3-7B, 3-10B
\\
\bottomrule
\end{tabular}
\end{table}

\paragraph{Transfer across model variants.}
We onboard ten additional checkpoints spanning 1.24B--14.8B
parameters without updating the interface
(Table~\ref{tab:onboarded-variants}).
All achieve minimum ID R@1 of 99.9--100.0\%.
Their behavioral directions also largely follow the geometry established
by the source models across scales and checkpoint variants
(Figure~\ref{fig:genrob}a).
Thus, the frozen interface transfers across scale and related
checkpoints with only isolated axis-specific deviations.

\paragraph{Calibration efficiency.}
For OLMo, R@1 rises from 88.4--89.5\% with 3K matched positions to
99.6\% with 10K, with no material improvement at 30K or 100K
(Figure~\ref{fig:genrob}b).
High-resolution onboarding therefore requires only one tenth of the
default 100K-position calibration pool.

\section{Conclusion}

We introduced a universal activation interface that connects compatible language models through one encoder--decoder adapter pair per model. Once frozen, new models join by training only their own adapter pair, without changing existing coordinates, paths, or tools. This model-wise design replaces pair-specific connectors with an extensible activation ecosystem that remains stable across compatible model families, scales, and calibration budgets.

\newpage

\appendix
\setcounter{secnumdepth}{2}

\bibliography{arxiv/arxiv}

\begin{table*}[t]
\centering
\caption{Model pool. Source models are jointly trained; OLMo-2-7B is added later
by \emph{freeze-then-onboard}; Qwen2.5-7B is the NLA carrier. The cut block
$\ell_m$ is at half depth for every model except the carrier ($\dagger$), using the official NLA hook at layer~20}
\label{tab:models}
\setlength{\tabcolsep}{6pt}
\begin{tabular}{l l l c r r r l}
\toprule
\textbf{Alias} & \textbf{Checkpoint} & \textbf{Org.} & \textbf{Base/Inst.} &
\textbf{Params} & \textbf{$d_m$} & \textbf{$L_m$ / $\ell_m$} & \textbf{Role} \\
\midrule
Llama-3.2-3B\footnotemark[1]      & Llama-3.2-3B-Instruct    & Meta    & Inst. & 3.2B & 3072 & 28 / 14 & source \\
Qwen3-4B\footnotemark[2]          & Qwen3-4B-Instruct-2507   & Alibaba & Inst. & 4.0B & 2560 & 36 / 18 & source \\
SmolLM2-1.7B\footnotemark[3]      & SmolLM2-1.7B-Instruct    & HF      & Inst. & 1.7B & 2048 & 24 / 12 & source \\
Falcon3-3B\footnotemark[4]        & Falcon3-3B-Base          & TII     & Base  & 3.1B & 3072 & 22 / 11 & source \\
OLMo-2-7B\footnotemark[5]         & OLMo-2-1124-7B           & AllenAI & Base  & 7.3B & 4096 & 32 / 16 & onboarded \\
Qwen2.5-7B\footnotemark[6]        & Qwen2.5-7B-Instruct      & Alibaba & Inst. & 7.6B & 3584 & 28 / 20$^{\dagger}$ & carrier (NLA) \\
\bottomrule
\end{tabular}
\end{table*}

\begin{table*}[t]
\centering
\small
\caption{HuggingFace sources for the nine behavioral axes. Positive/negative are
the two sides of each axis.}
\label{tab:axes}
\setlength{\tabcolsep}{10pt}
\begin{tabular}{l l l}
\toprule
\textbf{Axis} & \textbf{Positive source} & \textbf{Negative source} \\
\midrule
refusal    & \texttt{allenai/wildjailbreak} (harmful) & \texttt{allenai/wildjailbreak} (benign) \\
math       & \texttt{openai/gsm8k}                    & \texttt{sam-paech/mmlu-pro-nomath-sml} \\
science    & \texttt{Idavidrein/gpqa}                 & \texttt{TIGER-Lab/MMLU-Pro} (non-STEM) \\
factual    & \texttt{cais/mmlu}                       & \texttt{Anthropic/llm\_global\_opinions} \\
sycophancy & \texttt{meg-tong/sycophancy-eval} (are-you-sure) & \texttt{meg-tong/sycophancy-eval} (baseline) \\
toxicity   & \texttt{google/civil\_comments} ($>0.5$) & \texttt{ag\_news} \\
sentiment  & \texttt{stanfordnlp/sst2} (positive)     & \texttt{stanfordnlp/sst2} (negative) \\
emotion    & \texttt{dair-ai/emotion} (anger)         & \texttt{dair-ai/emotion} (joy) \\
bias       & \texttt{oskarvanderwal/bbq} (ambiguous)  & \texttt{oskarvanderwal/bbq} (disambiguated) \\
\bottomrule
\end{tabular}
\end{table*}

\newpage

\input{arxiv/appendix}

\end{document}

%% file: arxiv/appendix.tex
% ==========================================================================
% Appendix A — Detailed Setup and Interface Construction
% ==========================================================================

\footnotetext[1]{\url{https://huggingface.co/meta-llama/Llama-3.2-3B-Instruct}}
\footnotetext[2]{\url{https://huggingface.co/Qwen/Qwen3-4B-Instruct-2507}}
\footnotetext[3]{\url{https://huggingface.co/HuggingFaceTB/SmolLM2-1.7B-Instruct}}
\footnotetext[4]{\url{https://huggingface.co/tiiuae/Falcon3-3B-Base}}
\footnotetext[5]{\url{https://huggingface.co/allenai/OLMo-2-1124-7B}}
\footnotetext[6]{\url{https://huggingface.co/Qwen/Qwen2.5-7B-Instruct}}

\section{Detailed Setup and Interface Construction}
\label{app:setup}

This section specifies the models, data, activation extraction,
normalization, architecture, and optimization settings used to
reproduce the interface.

% --------------------------------------------------------------------------
\subsection{Models}
\label{app:models}

We use four \emph{source} models, OLMo-2-7B as the main
\emph{onboarded} model, and Qwen2.5-7B as the \emph{carrier} model for
the NLA experiment.
All LLM parameters remain frozen throughout.
For each model, we extract the residual-stream activation after a
selected transformer block, referred to as the \emph{cut} block.
The default cut is at normalized depth $0.5$,
$\ell_m=L_m/2$; the NLA carrier uses its released layer-20 interface.
Forward passes use \texttt{bfloat16}, and extracted activations are
stored in \texttt{float16}.
Table~\ref{tab:models} reports the details.

The ten additional checkpoints used for the scale and generality
experiments are listed in Appendix~\ref{app:generality}.

% --------------------------------------------------------------------------
\subsection{Interface Corpus and Evaluation Data}
\label{app:corpus}

\paragraph{Interface corpus.}
The interface corpus combines web text, instructions, code, and
mathematical problems.
Documents are restricted to 600--4,000 characters.
Table~\ref{tab:corpus} gives the corpus composition.

\begin{table}[t]
\centering
\caption{Interface corpus.
The corpus contains 118,563 documents and 1,422,735 matched
positions.}
\label{tab:corpus}
\setlength{\tabcolsep}{8pt}
\begin{tabular}{l r r}
\toprule
\textbf{Dataset} & \textbf{\# docs} & \textbf{\% docs} \\
\midrule
OpenWebText\footnotemark[7]   & 99{,}997 & 84.3 \\
Alpaca\footnotemark[8]        &  9{,}839 &  8.3 \\
CodeContests\footnotemark[9]  &  6{,}250 &  5.3 \\
GSM8K\footnotemark[10]        &  2{,}477 &  2.1 \\
\midrule
\textbf{Total}                & \textbf{118{,}563} & \textbf{100} \\
\bottomrule
\end{tabular}
\end{table}
\footnotetext[7]{\url{https://huggingface.co/datasets/Skylion007/openwebtext}}
\footnotetext[8]{\url{https://huggingface.co/datasets/tatsu-lab/alpaca}}
\footnotetext[9]{\url{https://huggingface.co/datasets/deepmind/code_contests}}
\footnotetext[10]{\url{https://huggingface.co/datasets/openai/gsm8k}}

\paragraph{Splits and evaluation pools.}
We split the corpus at the document level into 98\% training and 2\%
held-out documents.
The held-out split contains 27,804 matched positions, ensuring that
prefixes from the same document never appear in both splits.
The in-distribution retrieval pool contains 3,600 positions sampled
from the held-out documents while preserving the corpus mixture.
Default onboarding uses 100,000 calibration positions sampled from the
training split.

\paragraph{Behavioral evaluation axes.}
The nine behavioral axes in Table~\ref{tab:axes} are independent of the
interface corpus and are used only after the interface is frozen.
Each axis contains 100 positive and 100 negative texts for direction
and probe fitting, together with an equally sized disjoint test split.
This yields 400 texts per axis and 3,600 texts in total.
No behavioral text or label is used for interface training or
onboarding.

% --------------------------------------------------------------------------
\subsection{Matched-Prefix Construction Across Tokenizers}
\label{app:prefix}

To condition all source models on exactly the same raw text, we sample
candidate byte offsets and retain only endpoints that coincide with a
token boundary in all four source tokenizers.
The corresponding final-token activations therefore represent the same
text prefix despite differences in token segmentation.

We further measure leave-one-out boundary coverage in Table~\ref{tab:coverage}.
For each model, we construct endpoints using the other four main
models and report the fraction that are also exact boundaries for the
held-out tokenizer. Coverage is estimated over 3,000 documents.
Coverage exceeds 99.8\% for every evaluated model, so token-boundary
mismatches are rare.

\begin{table}[t]
\centering
\setlength{\tabcolsep}{10pt}
\caption{
Leave-one-out token-boundary coverage.
Each value is the fraction of endpoints shared by the other four main
models that are also valid boundaries for the held-out tokenizer.
}
\label{tab:coverage}
\begin{tabular}{@{}lc@{}}
\toprule
Held-out model & Coverage (\%) \\
\midrule
Llama-3.2-3B  & 100.00 \\
Qwen3-4B      & 100.00 \\
SmolLM2-1.7B  &  99.97 \\
Falcon3-3B    &  99.83 \\
OLMo-2-7B     & 100.00 \\
\midrule
Qwen2.5-7B carrier & 100.00 \\
\bottomrule
\end{tabular}
\end{table}

\subsection{Per-Model Normalization and Inverse Mapping}
\label{app:norm}

Each model uses a fixed normalization $N_m$ consisting of a
per-dimension z-score followed by regularized ZCA whitening.
The statistics are estimated once from approximately 200,000 training
activations in \texttt{float64} and remain frozen afterward.

\paragraph{Z-score.}
Let $\boldsymbol{\mu}_m$ and $\boldsymbol{\sigma}_m$ denote the
per-dimension mean and stabilized standard deviation,
with
$\boldsymbol{\sigma}_m
=\operatorname{std}(\mathbf{h}_m)+10^{-6}$.
The standardized activation is
\begin{equation}
\mathbf{u}_m(\mathbf{h})
=
(\mathbf{h}-\boldsymbol{\mu}_m)
\oslash
\boldsymbol{\sigma}_m ,
\label{eq:app-zscore}
\end{equation}
where $\oslash$ denotes element-wise division.

\paragraph{ZCA whitening.}
Let $\mathbf{X}_m\in\mathbb{R}^{N\times d_m}$ contain the standardized
training activations.
We compute
\begin{equation}
\begin{aligned}
\mathbf{C}_m
&=
\frac{1}{N}\mathbf{X}_m^\top\mathbf{X}_m
=
\mathbf{V}_m
\operatorname{diag}(\boldsymbol{\lambda}_m)
\mathbf{V}_m^\top,
\\
\widetilde{\boldsymbol{\lambda}}_m
&=
\max(\boldsymbol{\lambda}_m,0)
+
\rho\,\bar{\lambda}_m,
\qquad
\rho=10^{-2},
\\
\mathbf{W}_m
&=
\mathbf{V}_m
\operatorname{diag}
\left(
\widetilde{\boldsymbol{\lambda}}_m^{-1/2}
\right)
\mathbf{V}_m^\top ,
\end{aligned}
\label{eq:app-zca}
\end{equation}
where $\bar{\lambda}_m$ is the mean eigenvalue.
The additive regularization prevents low-variance directions from
receiving excessively large weights.
No dimensional truncation is applied.
The complete normalization is
\begin{equation}
N_m(\mathbf{h})
=
\mathbf{W}_m\mathbf{u}_m(\mathbf{h}).
\label{eq:app-normalizer}
\end{equation}

\paragraph{Inverse mapping.}
The inverse whitening matrix and native-space reconstruction are
\begin{equation}
\begin{aligned}
\mathbf{W}_m^{-1}
&=
\mathbf{V}_m
\operatorname{diag}
\left(
\widetilde{\boldsymbol{\lambda}}_m^{1/2}
\right)
\mathbf{V}_m^\top,
\\
N_m^{-1}(\mathbf{y})
&=
\left(
\mathbf{W}_m^{-1}\mathbf{y}
\right)
\odot\boldsymbol{\sigma}_m
+
\boldsymbol{\mu}_m .
\end{aligned}
\label{eq:app-inverse-normalizer}
\end{equation}
where $\odot$ denotes element-wise multiplication.
Reconstruction and onboarding losses are computed directly in the
normalized space using mean squared error.
The inverse is applied only when a native activation is required.

% --------------------------------------------------------------------------
\subsection{Architecture and Optimization}
\label{app:arch}

\paragraph{Architecture.}
Each model has a linear encoder adapter
$A_m^{\mathrm{enc}}:\mathbb{R}^{d_m}\rightarrow\mathbb{R}^{D}$
and decoder adapter
$A_m^{\mathrm{dec}}:\mathbb{R}^{D}\rightarrow\mathbb{R}^{d_m}$.
The shared modules $\Phi^{\mathrm{in}}$ and $\Phi^{\mathrm{out}}$
each contain two pre-normalized residual MLP blocks,
\begin{equation}
\operatorname{Block}(\mathbf{x})
=
\mathbf{x}
+
\mathbf{W}_2
\operatorname{GELU}
\left(
\mathbf{W}_1\operatorname{LN}(\mathbf{x})
\right),
\label{eq:app-residual-block}
\end{equation}
at shared width $D=3072$.
The model-specific adapters remain linear.
An encoder--decoder adapter pair contains 18.9M parameters when
$d_m=3072$, and 12.6M--25.2M across the main model widths.

\paragraph{Optimization.}
Table~\ref{tab:optim} summarizes the optimization settings.
Source training jointly optimizes self-reconstruction,
cross-reconstruction, and all-pairs InfoNCE.
Onboarding freezes the source interface and trains only the new adapter
pair using self-reconstruction and the fixed source-consensus target.

\begin{table}[t]
\centering
\small
\setlength{\tabcolsep}{7pt}
\caption{
Optimization settings for source-interface training and frozen
onboarding.
}
\label{tab:optim}
\begin{tabular}{@{}lll@{}}
\toprule
Setting & Source interface & Onboarding \\
\midrule
Optimizer
& AdamW
& AdamW
\\
Adapters LR
& $10^{-3}$
& $10^{-3}$
\\
Shared LR
& $3{\times}10^{-4}$
& --
\\
Scheduler
& cosine
& constant
\\
Batch size
& 4,096
& 4,096
\\
Steps
& 12,000
& 3,000
\\
Position pool
& training split
& calibration pool
\\
Precision
& bf16 autocast
& bf16 autocast
\\
Gradient clip
& 1.0
& none
\\
$\lambda_{\mathrm{self}}$
& 1.0
& 1.0
\\
$\lambda_{\mathrm{cross}}$
& 1.0
& --
\\
$\lambda_{\mathrm{nce}}$
& 0.1
& --
\\
$\lambda_{\mathrm{align}}$
& --
& 1.0
\\
Temperature $\tau$
& 0.07
& --
\\
Seed
& 0
& 0
\\
\bottomrule
\end{tabular}
\end{table}

\paragraph{Scaling.}
Each connected model adds one encoder--decoder adapter pair, while the
shared modules remain fixed.
Thus, persistent parameters grow linearly with the number of models,
although the current all-pairs source objective has quadratic training
cost.
Onboarding optimizes only the new adapter pair without retraining any
existing model or tool.

\begin{table}[t]
\centering

\setlength{\tabcolsep}{7pt}
\caption{
Label-free compatibility screen using Llama-3.2-3B as the reference
$r$ and each candidate as $m$.
R@1 measures held-out retrieval from ridge-predicted activations.
FVE reports reconstruction quality for the two directional maps.
Models with R@1 above 50\% are retained.
}
\label{tab:screen}
\begin{tabular}{@{}lccc@{}}
\toprule
Model
& R@1 (\%)
& FVE $r{\rightarrow}m\,/\,m{\rightarrow}r$
& Decision \\
\midrule
Qwen3-4B
& 98.6
& .400\,/\, .380
& Retain \\
SmolLM2-1.7B
& 99.6
& .408\,/\, .356
& Retain \\
Falcon3-3B
& 99.3
& .373\,/\, .373
& Retain \\
OLMo-2-7B
& 98.9
& .345\,/\, .444
& Retain \\
\midrule
Phi-3.5-mini
& 1.7
& .180\,/\, $-.030$
& Reject \\
Gemma-3-4B
& 1.0
& .100\,/\, $-.040$
& Reject \\
Mistral-7B
& 2.6
& .044\,/\, $-.052$
& Reject \\
Yi-1.5-6B
& 4.7
& $-.060$\,/\, $-.102$
& Reject \\
Apertus-8B
& 0.1
& .300\,/\, $-1.334$
& Reject \\
\bottomrule
\end{tabular}
\end{table}

\section{Compatibility Screening and Interface Validation}
\label{app:compatibility}

This section expands the intrinsic evaluation in
Section~\ref{sec:intrinsic}.
We report the label-free compatibility screen, shared-space retrieval,
round-trip fidelity, and behavioral geometry in the full shared space.

\subsection{Evaluation Metrics}
\label{app:intrinsic-metrics}

\paragraph{Cross-model retrieval.}
For a pool of $K$ matched inputs, retrieval from model $m$ to model $n$
uses each shared representation from $m$ as a query and searches among
the $K$ representations produced by $n$.
R@1 is the fraction of queries whose nearest representation by cosine
similarity corresponds to the same input:
\begin{equation}
\operatorname{R@1}_{m\rightarrow n}
=
\frac{1}{K}
\sum_{i=1}^{K}
\mathbbm{1}
\left[
i
=
\arg\max_{j}
\operatorname{cos}
\left(
\mathbf{z}_m(x_i),
\mathbf{z}_n(x_j)
\right)
\right].
\label{eq:retrieval-r1}
\end{equation}
Thus, high R@1 means that the two models assign nearby shared
representations to the same text, even among many competing inputs.

\paragraph{Round-trip fidelity.}
Self-FVE measures how well model $m$'s activation is reconstructed
after passing through its own encoder and decoder.
Let
$\overline{\mathbf{u}}_m$ be the mean normalized activation over the
evaluation set.
We compute
\begin{equation}
\operatorname{SelfFVE}_m
=
1-
\frac{
\sum_i
\operatorname{mse}
\left(
N_m(\widetilde{\mathbf{h}}_{m\leftarrow m}(x_i)),
N_m(\mathbf{h}_m(x_i))
\right)
}{
\sum_i
\operatorname{mse}
\left(
N_m(\mathbf{h}_m(x_i)),
\overline{\mathbf{u}}_m
\right)
}.
\label{eq:self-fve}
\end{equation}
A value of 1 indicates perfect reconstruction, 0 matches the
mean-activation baseline, and a negative value is worse than that
baseline.
All errors are measured in the model's normalized activation space.

% --------------------------------------------------------------------------
\subsection{Label-Free Compatibility Screening}
\label{app:screen}

Before bus training, we test whether each candidate model has a simple
linear correspondence with a fixed Llama-3.2-3B reference.
We fit closed-form ridge maps in both directions using 40,000 matched
positions, with $\lambda=100$, and evaluate them on 4,096 held-out
positions.
R@1 is computed from the ridge-predicted and target activations using
Equation~\eqref{eq:retrieval-r1}.
Models with R@1 above 50\% are retained.

We also report FVE as a diagnostic of linear reconstruction quality.
FVE measures the fraction of variance in the target activations
explained by the ridge prediction; negative values indicate performance
worse than predicting the target mean.
We use $r$ for the Llama reference and $m$ for the candidate model, so
$r{\rightarrow}m$ and $m{\rightarrow}r$ denote the two mapping
directions.

Across the full candidate screen, R@1 is sharply bimodal:
98.6--99.6\% for compatible models and 0.1--4.7\% for rejected
models, with no candidate near the 50\% threshold.
Replacing the Llama reference with Qwen preserves this partition.
FVE is more sensitive to the choice of reference, and two models cross
a fixed FVE threshold despite successful onboarding.

% --------------------------------------------------------------------------
\subsection{Reconstruction Fidelity}
\label{app:fidelity}

Retrieval measures whether corresponding inputs occupy nearby locations
in the shared space.
We separately evaluate how much native activation information remains
after encoding and decoding.
Table~\ref{tab:selffid} reports round-trip fidelity for the final
interface, which contains four source models and onboarded OLMo.
Self-FVE follows Equation~\eqref{eq:self-fve}.
Bus $\Delta$CE measures the change in next-token cross-entropy after
replacing the native cut-layer state with its encode--decode
reconstruction.

\begin{table}[t]
\centering
\setlength{\tabcolsep}{9pt}
\caption{
Round-trip fidelity of the final interface.
Self-FVE is evaluated on held-out activations, and
$\Delta\mathrm{CE}$ on held-out prefixes.
Higher self-FVE is better; negative $\Delta\mathrm{CE}$ indicates a
small improvement in next-token prediction.
}
\label{tab:selffid}
\begin{tabular}{@{}lccc@{}}
\toprule
Model & self-FVE & Bus $\Delta\mathrm{CE}$ & $d_m/D$ \\
\midrule
Llama-3.2-3B  & .982 & $-.010$ & 1.00 \\
Qwen3-4B      & .986 & $-.005$ & 0.83 \\
SmolLM2-1.7B  & .989 & $-.008$ & 0.67 \\
Falcon3-3B    & .983 & $-.002$ & 1.00 \\
OLMo-2-7B     & .734 & $+.043$ & 1.33 \\
\bottomrule
\end{tabular}
\end{table}

The four source models retain self-FVE of 0.982--0.989 with negligible
changes in next-token loss.
OLMo has lower self-FVE and a slightly larger $\Delta\mathrm{CE}$.
This pattern is consistent with its native width exceeding the shared
dimension, $d_m=4096>D=3072$.
It also shows that near-perfect shared-space retrieval does not require
exact round-trip reconstruction.

% --------------------------------------------------------------------------
\subsection{Full-Space Behavioral Geometry}
\label{app:geometry}

The main PCA provides a two-dimensional view of behavioral directions.
We verify the same structure in the full
$D=3072$ shared space.
Across nine axes and ten model pairs, the mean cosine similarity
between directions for the same axis is 0.815, compared with
$-0.014$ between different axes.
The resulting separation is
$\Delta_{\mathrm{axis}}=0.829$.
Pairs involving onboarded OLMo obtain same-axis cosine 0.807, close to
0.820 for source--source pairs.

% =========================================================

\begin{table*}[t]
\centering
\setlength{\tabcolsep}{5pt}
\caption{
Linear-probe data splits.
Only the four source models provide training examples; OLMo is never
used during probe fitting.
All test texts are disjoint from the training texts.
}
\label{tab:probe-protocol}
\begin{tabular}{@{}lccc@{}}
\toprule
Task & Training set across four sources & Test set per model & Metric \\
\midrule
Binary, per axis
& $4$ models $\times$ $200$ texts $=800$
& $200$ texts
& AUROC \\
Nine-way
& $4$ models $\times$ $9$ axes $\times$ $100$ texts $=3{,}600$
& $9$ axes $\times$ $100$ texts $=900$
& Accuracy \\
\bottomrule
\end{tabular}
\end{table*}

\begin{table*}[t]
\centering
\setlength{\tabcolsep}{6pt}
\caption{
Functional fidelity of the shared SAE.
The three middle columns report the increase in next-token
cross-entropy, in nats, after using the shared-SAE reconstruction,
native-SAE reconstruction, or a zero state.
Lower $\Delta\mathrm{CE}$ is better.
Utility retention places the native SAE at 100\% and the zero-state
ablation at 0\%.
OLMo$^\dagger$ is unseen during shared-SAE training.
}
\label{tab:sae-utility}
\begin{tabular}{@{}lrrrr@{}}
\toprule
Model
& \shortstack{Shared\\SAE}
& \shortstack{Native\\SAE}
& \shortstack{Zero\\state}
& \shortstack{Utility\\retained (\%)} \\
\midrule
Llama
& 0.268
& 0.090
& 4.823
& 96.2 \\
Qwen3
& 0.349
& 0.061
& 6.794
& 95.7 \\
SmolLM2
& 0.284
& 0.078
& 6.579
& 96.8 \\
Falcon3
& 0.189
& 0.082
& 8.682
& 98.8 \\
OLMo$^\dagger$
& 0.398
& 0.128
& 4.485
& 93.8 \\
\bottomrule
\end{tabular}
\end{table*}

\section{Activation-Tool Evaluation Details}
\label{app:tools}

This section provides the evaluation protocols and additional
diagnostics for the linear probes, shared SAE, and carrier-specific
NLA in Section~\ref{sec:tools}.
We report the classifier settings, SAE functional fidelity, NLA
reconstruction controls, and per-axis NLA performance omitted from the
main paper.

% --------------------------------------------------------------------------
\subsection{Shared Linear Probes}
\label{app:probe}

\paragraph{Protocol.}
The interface is frozen before probe training.
All shared states are converted to \texttt{float32} and unit-normalized,
with no additional standardization.
We fit $\ell_2$-regularized logistic regressions using
\texttt{scikit-learn} with \texttt{lbfgs}, $C=1$,
\texttt{max\_iter}=2000, and seed 42.
Classes are balanced by construction, so we use neither class
weighting nor threshold tuning.
The native-space reference probes use the same settings.

The pooled probes in Table~\ref{tab:probe} are trained once on the
source models and then applied unchanged to each model.
Their binary performance closely matches that of probes fitted
separately in the native activation spaces.
OLMo therefore tests whether the same decision boundaries remain usable
after onboarding, without model-specific probe training or calibration.

% --------------------------------------------------------------------------
% --------------------------------------------------------------------------
\subsection{Shared Sparse Autoencoder}
\label{app:sae}

\paragraph{Training setup.}
We train one TopK SAE~\citep{gao2024scaling} on 400,000
model-balanced shared states from Llama, Qwen3, SmolLM2, and Falcon3.
The dictionary contains 32,768 features, with $k=32$ active features
per input and auxiliary $k=256$ for dead-feature recovery.
Training uses Adam with learning rate $10^{-4}$ and 12,000 steps.
OLMo is excluded from SAE training and is evaluated only through its
onboarding adapter.
For comparison, we independently train one native SAE for each model
using the same dictionary size, sparsity, data volume, and optimization
settings.

\paragraph{Reconstruction and functional utility.}
Fraction of variance unexplained (FVU) measures the SAE reconstruction
error relative to the variance of its input states; lower values are
better.
The shared SAE obtains FVU of 0.132--0.145 on the four source models
and 0.106 on OLMo, with no dead features.
Because these values are all measured in the same shared space, they
can be compared across models.

We additionally test whether the reconstructed activation remains
useful to the model.
We replace the cut-layer state with its SAE reconstruction and measure
the resulting increase in next-token cross-entropy,
$\Delta\mathrm{CE}$.
Here, \emph{shared} denotes the single SAE trained in the shared space,
\emph{native} denotes an SAE trained separately in the model's own
activation space, and \emph{zero} replaces the activation with a zero
state.
For model $m$, utility retention is
\begin{equation}
R_m^{\mathrm{SAE}}
=
\frac{
\Delta\mathrm{CE}^{\mathrm{zero}}_m
-
\Delta\mathrm{CE}^{\mathrm{shared}}_m
}{
\Delta\mathrm{CE}^{\mathrm{zero}}_m
-
\Delta\mathrm{CE}^{\mathrm{native}}_m
}.
\label{eq:sae-retention}
\end{equation}
A value of 1 matches the native SAE, while 0 matches the zero-state
ablation.

As shown in Table~\ref{tab:sae-utility}, the shared SAE causes a larger raw $\Delta\mathrm{CE}$ than the
separately trained native SAEs, but retains 93.8--98.8\% of their
utility relative to the zero-state baseline.
This modest loss provides one dictionary and one set of feature indices
for every connected model, including OLMo without SAE retraining.

\paragraph{Feature overlap and example selection.}
Table~\ref{tab:sae-jaccard} measures how often two models activate the
same Top-$32$ feature indices on matched inputs.
We additionally examine individual features by comparing their
top-activating input sets across models.
Among the 590 candidate features in this analysis, the median
feature-level Jaccard is 0.569 and the 90th percentile is 0.778.
The qualitative features in Table~\ref{tab:sae-nla-demos} were selected
from this high-overlap group before their text examples were inspected.
They are therefore illustrative high-agreement features rather than
random samples from the full dictionary.

\begin{table*}[t]
\centering
\setlength{\tabcolsep}{7pt}
\caption{
NLA reconstruction fidelity.
Cosine compares the AR reconstruction with the true carrier activation
for the same input; mismatch uses an activation from a different input.
Retention sets the native-carrier result to 100\% and the mismatch
control to 0\%.
}
\label{tab:nla-fidelity}
\begin{tabular}{@{}lccc@{}}
\toprule
Activation source
& Cosine
& Mismatch
& Retention (\%) \\
\midrule
Qwen2.5 carrier
& 0.866
& 0.328
& 100.0 \\ \midrule
Llama
& 0.827
& 0.326
& 93.1 \\
Qwen3
& 0.820
& 0.328
& 91.6 \\
SmolLM2
& 0.824
& 0.329
& 92.1 \\
Falcon3
& 0.820
& 0.327
& 91.8 \\
OLMo$^\dagger$
& 0.848
& 0.331
& 96.1 \\
\bottomrule
\end{tabular}
\end{table*}

\begin{table*}[t]
\centering
\small
\setlength{\tabcolsep}{6pt}
\caption{
Per-axis recall (\%) of the nine-way classifier trained only on native
carrier verbalizations.
Routed sources give the minimum--maximum over Llama, Qwen3, SmolLM2,
and Falcon3.
OLMo$^\dagger$ is evaluated after adapter-only onboarding.
}
\label{tab:nla-axis-recall}
\begin{tabular}{@{}lccc@{}}
\toprule
Axis
& Native carrier
& Routed sources
& OLMo$^\dagger$ \\
\midrule
Refusal
& 96
& 74--88
& 90 \\
Math
& 98
& 88--98
& 100 \\
Science
& 92
& 76--90
& 86 \\
Factual
& 94
& 88--100
& 92 \\
Sycophancy
& 100
& 100
& 100 \\
Toxicity
& 96
& 84--94
& 96 \\
Sentiment
& 100
& 92--98
& 6 \\
Emotion
& 86
& 86--92
& 96 \\
Bias
& 96
& 90--100
& 98 \\
\bottomrule
\end{tabular}
\end{table*}

% --------------------------------------------------------------------------
\subsection{NLA: Carrier Routing and Evaluation}
\label{app:nla}

\paragraph{Carrier and routing setup.}
We use the released activation verbalizer~(AV) and activation
reconstructor~(AR) for Qwen2.5-7B at layer 20, without modifying their
weights, prompts, or inference settings.
The Qwen2.5 carrier is attached to the frozen interface using 100,000
unlabeled calibration positions and 3,000 adapter-only optimization
steps.
It reaches 100\% ID R@1 and self-FVE of 0.833.
To process an activation from another model, the interface first
decodes the shared state into the Qwen2.5 activation space, after which
the released AV and AR are applied unchanged.
The NLA itself receives no additional training.

\paragraph{Reconstruction fidelity.}
We evaluate whether an NLA verbalization retains enough information to
reconstruct the corresponding carrier activation.
Cosine similarity is measured between the AR reconstruction and the
true Qwen2.5 activation produced by the same input.
As a control, \emph{mismatch} compares the reconstruction with a
carrier activation from a different input.
Retention rescales each routed cosine between the mismatch floor
(0\%) and the native-carrier result (100\%).

All routed models remain well above the mismatched-input control and
retain 91.6--96.1\% of the native reconstruction signal.
OLMo obtains the highest routed cosine despite being unseen during both
source-interface and NLA training.
Thus, carrier routing preserves substantial input-specific information
even when the source activation comes from an onboarded model.

\paragraph{Behavioral information in verbalizations.}
We next test whether the verbalizations preserve the behavioral axis of
the input.
A TF--IDF classifier with unigram and bigram features and an
$\ell_2$-regularized logistic regression is trained only on native
Qwen2.5 verbalizations.
The training split contains 60 examples per axis and the test split
contains 50 per axis.
The same classifier is then applied unchanged to verbalizations routed
from every other model.
Table~\ref{tab:nla-axis-recall} reports recall for each axis.

The cross-axis control, which deliberately mismatches outputs and axis
labels, reaches only 0.7\% accuracy, compared with chance accuracy of
11.1\%.
OLMo's lower overall accuracy is concentrated in sentiment, where
recall falls to 6\%; recall on the other eight axes remains
86--100\%.
The direct shared-space probe successfully identifies OLMo sentiment,
suggesting that this localized failure occurs during carrier routing or
verbalization rather than in the shared representation itself.

For reference, the same classifier trained and evaluated on the
original input texts reaches 79.6\% accuracy.
The higher separability of the verbalizations suggests that the AV
expresses behavioral cues in a more consistent vocabulary; it does not
imply that the NLA introduces new information.

\paragraph{Additional qualitative outputs.}
Table~\ref{tab:sae-nla-demos} presents one medical example.
Additional refusal and emotion examples show similar agreement in topic,
format, and overall framing, although named entities and specific
details sometimes differ.

% ===================================================
\begin{table*}[t]
\centering
\setlength{\tabcolsep}{4.2pt}
\renewcommand{\arraystretch}{1.06}
\caption{
Complete front--back execution results over 1,000 held-out prefixes
per direction.
Native $A$--$B$ is the agreement between the two native models.
Composed--$B$ and composed--$A$ compare the composed prediction with
the native back and front models, respectively.
Gain is composed--$B$ minus native $A$--$B$ agreement, in percentage
points, and $\Delta$CE is measured against native $B$ in nats.
L, Q, S, F, and O denote Llama, Qwen3, SmolLM2, Falcon3, and OLMo.
}
\label{tab:compose-full}
\begin{tabular}{@{}llrrrrr@{}}
\toprule
Front
& Back
& \shortstack{Native $A$--$B$ (\%)}
& \shortstack{Composed --$B$ (\%)}
& \shortstack{Gain (\%p)}
& \shortstack{Composed --$A$ (\%)}
& $\Delta$CE \\
\midrule
L & L & 100.0 & 96.9 & $-3.1$ & 96.9 & $-0.008$ \\
Q & L & 67.2  & 78.6 & $+11.4$ & 67.4 & $+0.010$ \\
S & L & 62.0  & 77.7 & $+15.7$ & 62.8 & $+0.068$ \\
F & L & 60.5  & 78.5 & $+18.0$ & 61.9 & $+0.071$ \\
O & L & 66.8  & 79.0 & $+12.2$ & 68.4 & $-0.023$ \\
\midrule
L & Q & 67.2  & 86.6 & $+19.4$ & 67.8 & $+0.047$ \\
Q & Q & 100.0 & 97.8 & $-2.2$ & 97.8 & $-0.008$ \\
S & Q & 62.2  & 85.4 & $+23.2$ & 62.8 & $+0.019$ \\
F & Q & 60.4  & 85.6 & $+25.2$ & 60.3 & $+0.059$ \\
O & Q & 66.9  & 86.2 & $+19.3$ & 66.8 & $+0.012$ \\
\midrule
L & S & 62.0  & 82.9 & $+20.9$ & 64.7 & $+0.050$ \\
Q & S & 62.2  & 83.6 & $+21.4$ & 63.3 & $+0.034$ \\
S & S & 100.0 & 96.6 & $-3.4$ & 96.6 & $-0.007$ \\
F & S & 64.4  & 84.3 & $+19.9$ & 65.3 & $+0.071$ \\
O & S & 64.4  & 81.6 & $+17.2$ & 64.3 & $+0.037$ \\
\midrule
L & F & 60.5  & 87.7 & $+27.2$ & 60.3 & $-0.008$ \\
Q & F & 60.4  & 85.7 & $+25.3$ & 59.8 & $-0.018$ \\
S & F & 64.4  & 84.5 & $+20.1$ & 64.8 & $-0.006$ \\
F & F & 100.0 & 97.5 & $-2.5$ & 97.5 & $-0.002$ \\
O & F & 58.7  & 84.8 & $+26.1$ & 60.8 & $-0.023$ \\
\midrule
L & O & 66.8  & 76.5 &  $+9.7$ & 66.2 & $+0.306$ \\
Q & O & 66.9  & 74.4 &  $+7.5$ & 65.4 & $+0.306$ \\
S & O & 64.4  & 78.0 & $+13.6$ & 63.9 & $+0.273$ \\
F & O & 58.7  & 72.4 & $+13.7$ & 61.7 & $+0.381$ \\
O & O & 100.0 & 92.4 & $-7.6$ & 92.4 & $+0.045$ \\
\bottomrule
\end{tabular}
\end{table*}

\newpage

\section{Executing Translated States: Full Results}
\label{app:composition}

This section provides the complete state-execution results summarized
in Section~\ref{sec:execution}.
All experiments use the frozen four-source interface and the onboarded
OLMo adapter, without execution-specific or direction-specific
training.

\paragraph{Protocol and metrics.}
For a direction $A\!\rightarrow\!B$, model $A$ processes a prefix up to
its selected cut layer.
We translate its final-position state through
$D_B(E_A(\mathbf{h}_A))$ and insert it at the corresponding position
in model $B$.
All other context states remain those produced by $B$, whose frozen
upper layers then predict the next token.
This setting evaluates matched-context state execution rather than
full-sequence model stitching.

Each direction uses the same 1,000 held-out prefixes.
We report agreement between the two native models, agreement of the
composed prediction with native $B$ and native $A$, and the
cross-entropy change
$\Delta\mathrm{CE}
=\mathrm{CE}_{A\rightarrow B}-\mathrm{CE}_{B}$.
Back-model agreement uses token identity in $B$'s vocabulary, while
cross-model comparisons use exact decoded-string equality.

\newpage

\paragraph{Results.}
Self paths retain 92.4--97.8\% agreement with their native
predictions.
Across all cross-model directions, composed predictions agree with
native $B$ on 72.4--87.7\%, exceeding the native $A$--$B$ agreement.
Agreement with native $A$ remains close to the original
$A$--$B$ baseline.
Thus, the translated state influences the computation, while the
receiving model primarily determines the next-token behavior.

Paths into Llama, Qwen3, SmolLM2, and Falcon3 incur only
$-0.023$ to $+0.071$ nats of additional cross-entropy.
OLMo is a more difficult receiver, with penalties of
$+0.273$ to $+0.381$ nats, consistent with its wider native space and
lower round-trip fidelity.

The examples in Table~\ref{tab:compose-demo} show the same
overall pattern.
Identity and parametric recall often follow the receiving model when
the native models disagree, while neutral facts are generally
preserved.
However, hallucinated affiliations and the contradictory
Qwen$\rightarrow$Llama identity response show that the operation is not
a complete model replacement.
Because only the final-position state is replaced and the receiver's
remaining context and KV cache stay unchanged, these results support
matched-context state execution rather than full-sequence model
stitching.

\begin{table*}[t]
\centering
\setlength{\tabcolsep}{3.7pt}
\caption{
Models attached to the frozen four-source interface.
OLMo is the main onboarded model, Qwen2.5-7B is the NLA carrier, and
the remaining rows are scale or tuning variants.
Self-FVE measures round-trip reconstruction in the onboarded model's
normalized activation space.
``Correct axes'' counts the positive-example sets whose majority is
assigned to the matching source-axis centroid; it is complementary to,
but distinct from, the direction PCA in the main paper.
}
\label{tab:additional-models}
\begin{tabularx}{\textwidth}{
@{}
l
l
l
r
r
c
@{}
}
\toprule
Role
& Model
& Hugging Face checkpoint
& Params (B)
& self-FVE
& Correct axes \\
\midrule
Variant
& Llama-3.2-1B
& \texttt{meta-llama/Llama-3.2-1B}
& 1.24
& .9800
& 9/9 \\

Variant
& Falcon3-1B
& \texttt{tiiuae/Falcon3-1B-Base}
& 1.67
& .9417
& 9/9 \\

Variant
& Qwen3-1.7B
& \texttt{Qwen/Qwen3-1.7B}
& 1.72
& .9782
& 9/9 \\

Variant
& SmolLM3-3B
& \texttt{HuggingFaceTB/SmolLM3-3B}
& 3.08
& .9791
& 9/9 \\

Main
& OLMo-2-7B
& \texttt{allenai/OLMo-2-1124-7B}
& 7.30
& .7343
& 9/9 \\

Variant
& Falcon3-7B
& \texttt{tiiuae/Falcon3-7B-Base}
& 7.46
& .9663
& 9/9 \\

Carrier
& Qwen2.5-7B
& \texttt{Qwen/Qwen2.5-7B-Instruct}
& 7.62
& .8326
& 9/9 \\

Variant
& Llama-3.1-8B
& \texttt{meta-llama/Llama-3.1-8B-Instruct}
& 8.03
& .7349
& 9/9 \\

Variant
& Qwen3-8B
& \texttt{Qwen/Qwen3-8B}
& 8.19
& .7312
& 9/9 \\

Variant
& Falcon3-10B
& \texttt{tiiuae/Falcon3-10B-Base}
& 10.30
& .9654
& 9/9 \\

Variant
& Qwen3-14B
& \texttt{Qwen/Qwen3-14B}
& 14.80
& .5909
& 9/9 \\
\bottomrule
\end{tabularx}
\end{table*}

\section{Generality, Robustness, and Sensitivity}
\label{app:generality}

This section tests whether the frozen four-source interface extends
beyond the main five-model setting.
We evaluate OLMo together with ten additional checkpoints and summarize
existing sensitivity results for calibration size, source-model count,
cut depth, and random seed.
Because shared-space alignment and native-state reconstruction are
distinct, we report behavioral-axis assignment and self-FVE separately.

% --------------------------------------------------------------------------
\subsection{Additional Onboarded Checkpoints}
\label{app:additional-models}

We attach OLMo and ten additional checkpoints to the same frozen
Llama--Qwen3--SmolLM2--Falcon3 interface.
Unless noted otherwise, each model uses 100,000 unlabeled calibration
positions and 3,000 adapter-only optimization steps.
No existing interface parameter is updated.

\paragraph{Behavioral geometry.}
For each onboarded model and behavioral axis, we encode the positive
examples into the shared space and assign them to the nearest
source-axis centroid by cosine similarity.
A model--axis case is correct when the majority of its examples select
the matching axis.
All representations and centroids are unit-normalized before this
comparison.
All 99 cases are correct: each of the eleven onboarded models aligns
with all nine source-defined behavioral axes.
The weakest individual accuracies occur for science/reasoning and bias,
but no axis changes its majority assignment.

\paragraph{Scale and reconstruction fidelity.}
Round-trip self-FVE ranges from 0.591 to 0.980 across the onboarded
checkpoints.
Several models with wider native activation spaces obtain lower
self-FVE, which is consistent with compression through the fixed
$D=3072$ shared space.
However, model family and architecture also vary, so we interpret this
as a descriptive trend rather than an isolated width effect.
Shared behavioral-axis assignment remains correct for every model in
Table~\ref{tab:additional-models}.

\subsection{Additional Sensitivity Checks}
\label{app:additional-sensitivity}

\paragraph{Calibration size.}
With 3K, 10K, 30K, and 100K unique calibration positions, OLMo
self-FVE is $-0.550$, 0.696, 0.722, and 0.734, respectively.
The 3K condition also uses a smaller batch size and should therefore be
viewed as a low-budget setting rather than a controlled size-only
comparison.
Most of the available round-trip fidelity is recovered with 10K
positions, followed by smaller gains at 30K and 100K.

\paragraph{Source-model count.}
For the nested source sets Llama--Qwen3,
Llama--Qwen3--SmolLM2, and
Llama--Qwen3--SmolLM2--Falcon3, OLMo self-FVE is
0.712, 0.723, and 0.734.
This result shows a modest improvement as source diversity increases.
Because only one nested source set is evaluated at each count, it is
not an average over all possible source combinations.

\paragraph{Cut depth and random seed.}
Across normalized cut depths from 0.25 to 0.75, OLMo self-FVE varies
smoothly from 0.771 to 0.762, indicating that the default depth 0.5 is
not a narrow optimum.
These runs use a separate data subsample, so their absolute FVE values
should be compared only within the sweep.
Three complete source-interface runs give ID R@1 of
$99.949\pm0.008$\% and mean self-FVE of
$0.9851\pm0.0001$, showing low sensitivity to the random seed.

%% file: arxiv.bbl
\begin{thebibliography}{23}
\providecommand{\natexlab}[1]{#1}

\bibitem[{Achara et~al.(2026)Achara, Gaintseva, Mahaut, Chakraborty, Johansson, Barsbey, Rodol{\`a}, and Crisostomi}]{achara2026multiway}
Achara, A.; Gaintseva, T.; Mahaut, M.; Chakraborty, P.; Johansson, V.~S.; Barsbey, M.; Rodol{\`a}, E.; and Crisostomi, D. 2026.
\newblock Multi-Way Representation Alignment.
\newblock arXiv:2602.06205.

\bibitem[{Bansal, Nakkiran, and Barak(2021)}]{bansal2021stitching}
Bansal, Y.; Nakkiran, P.; and Barak, B. 2021.
\newblock Revisiting Model Stitching to Compare Neural Representations.
\newblock In \emph{Advances in Neural Information Processing Systems (NeurIPS)}.

\bibitem[{Chen et~al.(2025)Chen, Merullo, Stolfo, and Pavlick}]{chen2025featurestitching}
Chen, A.; Merullo, J.; Stolfo, A.; and Pavlick, E. 2025.
\newblock Transferring Linear Features Across Language Models With Model Stitching.
\newblock In \emph{Advances in Neural Information Processing Systems (NeurIPS)}.

\bibitem[{Fraser-Taliente et~al.(2026)Fraser-Taliente, Kantamneni, Ong, Mossing, Lu, Bogdan, Ameisen, Chen, Kishylau, Pearce, Tarng, Wu, Wu, Zhang, Ziegler, Hubinger, Batson, Lindsey, Zimmerman, and Marks}]{anthropic2026nla}
Fraser-Taliente, K.; Kantamneni, S.; Ong, E.; Mossing, D.; Lu, C.; Bogdan, P.~C.; Ameisen, E.; Chen, J.; Kishylau, D.; Pearce, A.; Tarng, J.; Wu, A.; Wu, J.; Zhang, Y.; Ziegler, D.~M.; Hubinger, E.; Batson, J.; Lindsey, J.; Zimmerman, S.; and Marks, S. 2026.
\newblock Natural Language Autoencoders Produce Unsupervised Explanations of {LLM} Activations.
\newblock \url{https://transformer-circuits.pub/2026/nla/index.html}.
\newblock Accessed: 2026-07-29.

\bibitem[{Gao et~al.(2025)Gao, Dupre~la Tour, Tillman, Goh, Troll, Radford, Sutskever, Leike, and Wu}]{gao2024scaling}
Gao, L.; Dupre~la Tour, T.; Tillman, H.; Goh, G.; Troll, R.; Radford, A.; Sutskever, I.; Leike, J.; and Wu, J. 2025.
\newblock Scaling and Evaluating Sparse Autoencoders.
\newblock In \emph{International Conference on Learning Representations (ICLR)}.

\bibitem[{Gorbett and Jana(2026)}]{gorbett2026characterizinglinearalignmentlanguage}
Gorbett, M.; and Jana, S. 2026.
\newblock Characterizing Linear Alignment Across Language Models.
\newblock arXiv:2603.18908.

\bibitem[{Hu et~al.(2025)Hu, Li, Agarwal, Lee, Jajoo, Li, Xu, Kim, Kim, Xu, Zhang, and Akella}]{hu2025stitchllm}
Hu, B.; Li, S.; Agarwal, S.; Lee, M.; Jajoo, A.; Li, J.; Xu, L.; Kim, G.-W.; Kim, D.; Xu, H.; Zhang, A.; and Akella, A. 2025.
\newblock StitchLLM: Serving LLMs, One Block at a Time.
\newblock In \emph{Proceedings of the 63rd Annual Meeting of the Association for Computational Linguistics (Volume 1: Long Papers)}, 26887--26903.

\bibitem[{Huh et~al.(2024)Huh, Cheung, Wang, and Isola}]{huh2024platonic}
Huh, M.; Cheung, B.; Wang, T.; and Isola, P. 2024.
\newblock Position: The Platonic Representation Hypothesis.
\newblock In \emph{Proceedings of the International Conference on Machine Learning (ICML)}.

\bibitem[{Jha et~al.(2025)Jha, Zhang, Shmatikov, and Morris}]{jha2025vec2vec}
Jha, R.; Zhang, C.; Shmatikov, V.; and Morris, J. 2025.
\newblock Harnessing the Universal Geometry of Embeddings.
\newblock In \emph{Advances in Neural Information Processing Systems (NeurIPS)}.

\bibitem[{Kessy, Lewin, and Strimmer(2018)}]{zca_whitening}
Kessy, A.; Lewin, A.; and Strimmer, K. 2018.
\newblock Optimal Whitening and Decorrelation.
\newblock \emph{The American Statistician}, 72(4): 309--314.

\bibitem[{Kim and Han(2026)}]{kim2026acs}
Kim, S.-H.; and Han, Y.-S. 2026.
\newblock Cross-Family Universality of Behavioral Axes via Anchor-Projected Representations.
\newblock arXiv:2605.09875.

\bibitem[{Kornblith et~al.(2019)Kornblith, Norouzi, Lee, and Hinton}]{kornblith2019cka}
Kornblith, S.; Norouzi, M.; Lee, H.; and Hinton, G. 2019.
\newblock Similarity of Neural Network Representations Revisited.
\newblock In \emph{Proceedings of the International Conference on Machine Learning (ICML)}.

\bibitem[{Liu et~al.(2026)Liu, Zhang, Yu, Xiong, He, Wu, Jung, Fredrikson, Wang, and Gao}]{liu2026visionwormhole}
Liu, X.; Zhang, R.; Yu, W.; Xiong, S.; He, L.; Wu, F.; Jung, H.; Fredrikson, M.; Wang, X.; and Gao, J. 2026.
\newblock The Vision Wormhole: Latent-Space Communication in Heterogeneous Multi-Agent Systems.
\newblock arXiv:2602.15382.

\bibitem[{Moschella et~al.(2023)Moschella, Maiorca, Fumero, Norelli, Locatello, and Rodol{\`a}}]{moschella2023relative}
Moschella, L.; Maiorca, V.; Fumero, M.; Norelli, A.; Locatello, F.; and Rodol{\`a}, E. 2023.
\newblock Relative Representations Enable Zero-Shot Latent Space Communication.
\newblock In \emph{International Conference on Learning Representations (ICLR)}.

\bibitem[{Nasiri-Sarvi, Rivaz, and Hosseini(2026)}]{nasiri2025sparc}
Nasiri-Sarvi, A.; Rivaz, H.; and Hosseini, M.~S. 2026.
\newblock {SPARC}: Concept-Aligned Sparse Autoencoders for Cross-Model and Cross-Modal Interpretability.
\newblock \emph{Transactions on Machine Learning Research}.

\bibitem[{Oozeer et~al.(2025)Oozeer, Nathawani, Prakash, Lan, Harrasse, and Abdullah}]{oozeer2025interventions}
Oozeer, N.; Nathawani, D.; Prakash, N.; Lan, M.; Harrasse, A.; and Abdullah, A. 2025.
\newblock Activation Space Interventions Can Be Transferred Between Large Language Models.
\newblock In \emph{Proceedings of the International Conference on Machine Learning (ICML)}.

\bibitem[{Puri et~al.(2025)Puri, Berend, Lapuschkin, and Samek}]{puri2025atlas}
Puri, B.; Berend, J.; Lapuschkin, S.; and Samek, W. 2025.
\newblock Atlas-Alignment: Making Interpretability Transferable Across Language Models.
\newblock arXiv:2510.27413.

\bibitem[{{Qwen Team}(2024)}]{qwen25}
{Qwen Team}. 2024.
\newblock Qwen2.5 Technical Report.
\newblock arXiv:2412.15115.

\bibitem[{Rimsky et~al.(2024)Rimsky, Gabrieli, Schulz, Tong, Hubinger, and Turner}]{panickssery2024caa}
Rimsky, N.; Gabrieli, N.; Schulz, J.; Tong, M.; Hubinger, E.; and Turner, A. 2024.
\newblock Steering Llama 2 via Contrastive Activation Addition.
\newblock In \emph{Proceedings of the 62nd Annual Meeting of the Association for Computational Linguistics (Volume 1: Long Papers)}, 15504--15522.

\bibitem[{Thasarathan et~al.(2025)Thasarathan, Forsyth, Fel, Kowal, and Derpanis}]{thasarathan2025usae}
Thasarathan, H.; Forsyth, J.; Fel, T.; Kowal, M.; and Derpanis, K. 2025.
\newblock Universal Sparse Autoencoders: Interpretable Cross-Model Concept Alignment.
\newblock In \emph{Proceedings of the 42nd International Conference on Machine Learning (ICML)}.

\bibitem[{van~den Oord, Li, and Vinyals(2019)}]{oord2018infonce}
van~den Oord, A.; Li, Y.; and Vinyals, O. 2019.
\newblock Representation Learning with Contrastive Predictive Coding.
\newblock arXiv:1807.03748.

\bibitem[{Zhang and Xin(2026)}]{zhang2026negative}
Zhang, P.; and Xin, J. 2026.
\newblock A Negative Result on Cross-Model Activation Transfer in a Pythia Multi-Hop Setting.
\newblock arXiv:2606.03280.

\bibitem[{Zhao et~al.(2026)Zhao, He, Wang, Payani, Li, and Du}]{zhao2026uav}
Zhao, H.; He, Z.; Wang, G.; Payani, A.; Li, Y.; and Du, M. 2026.
\newblock Universal Activation Verbalizer: A Unified Framework for Cross-Model Activation Explanation.
\newblock arXiv:2605.25903.

\end{thebibliography}
